\documentclass[10pt,final,3p]{elsarticle}
\usepackage{lmodern}

\usepackage{graphics}
\usepackage{graphicx}

\usepackage{booktabs}
\usepackage{multirow}
\usepackage{array}
\usepackage{amsmath}
\usepackage{amsthm}
\usepackage{verbatim}
\usepackage{relsize}
\usepackage[dvipsnames]{xcolor}
\usepackage{array}
\usepackage{bm}
\usepackage{caption}

\usepackage{algorithm}
\usepackage{algpseudocode}
\usepackage{enumitem}

\usepackage{natbib}
\usepackage{hyperref}
\newcommand{\doi}[1]{\textsc{doi}: \href{http://dx.doi.org/#1}{\nolinkurl{#1}}}

\definecolor{comment}{rgb}{0.0, 0.5, 0.0}

\usepackage{hyperref}

\hypersetup{
    colorlinks=true,
    linkcolor=blue,
    filecolor=magenta,
    urlcolor=cyan,
}
\usepackage{booktabs}
\usepackage{xcolor}
\usepackage{color}

\definecolor{mygreen}{rgb}{0,0.6,0}
\definecolor{mygray}{rgb}{0.5,0.5,0.5}
\definecolor{mymauve}{rgb}{0.58,0,0.82}

\usepackage{pdfcomment}

\usepackage{array}
\usepackage{mathtools}
\usepackage{wasysym}
\usepackage{verbatim}
\usepackage{relsize}
\usepackage{multirow}
\usepackage{multicol}

\usepackage{listings}
\usepackage[bitstream-charter]{mathdesign}
\usepackage[T1]{fontenc}
\usepackage[utf8]{inputenc} 

\usepackage[english]{babel}
\usepackage[normalem]{ulem}

\DeclareMathAlphabet{\pazocal}{OMS}{zplm}{m}{n}

\def \A {\pazocal{A}}

\usepackage{lscape}		
\usepackage{fancyvrb}

\usepackage[mathscr]{euscript}

\usepackage{textcomp}

\def \balpha {{\pmb{\alpha}}}
\def \bbeta  {{\pmb{\beta}}}

\def \bzero  {{\pmb{0}}}

\def \bPsi {{\boldsymbol\Psi}}
\def \xxi  {{\boldsymbol{\xi}}}

\def \X  {\pmb{X}} 
\def \x  {\pmb{x}} 

\def \cand {\mathrm{c}}
\def \neigh {\mathrm{s}}

\usepackage{bigints}

\usepackage[switch]{lineno}

  \usepackage{lineno}

\newcommand{\ns}{{\ensuremath{n_{\mathrm{sim}}}}}
\newcommand{\Ns}{{\ensuremath{N_{\mathrm{sim}}}}}

\newcommand{\xx}{\mbox{\boldmath $x$}}

\def \PCE {\textsf{PCE}}
\def \DALPCE {\textsf{DAL-PCE}}

\def \CMAME {{Computer  Methods in Applied Mechanics and Engineering}}

\journal{\CMAME}  

\begin{document}

\begin{frontmatter}



\title{Active Learning-based Domain Adaptive Localized Polynomial Chaos Expansion}


\author{Luk{\'a}{\v s} Nov{\'a}k\corref{cor1}}  \ead{novak.l@fce.vutbr.cz}  \cortext[cor1]{Corresponding author}
\address{Brno University of Technology, Brno, Czech Republic}
\author{Michael D. Shields}                     \ead{michael.shields@jhu.edu}
\address{Johns Hopkins University, Baltimore, USA}
\author{V{\'a}clav Sad{\'i}lek}                 \ead{sadilek.v@fce.vutbr.cz}

\author{Miroslav Vo{\v r}echovsk{\'y}}          \ead{vorechovsky.m@vut.cz}
\address{Brno University of Technology, Brno, Czech Republic}

\begin{abstract}
The paper presents a~novel methodology to build surrogate models of complicated functions by an active learning-based sequential decomposition of the input random space and construction of localized polynomial chaos expansions, referred to as domain adaptive localized polynomial chaos expansion (\DALPCE). The approach utilizes sequential decomposition of the input random space into smaller sub-domains approximated by low-order polynomial expansions. This allows approximation of functions with strong nonlinearties, discontinuities, and/or singularities. Decomposition of the input random space and local approximations alleviates the Gibbs phenomenon for these types of problems and confines error to a~very small vicinity near the non-linearity. The global behavior of the surrogate model is therefore significantly better than existing methods as shown in numerical examples. The whole process is driven by an active learning routine that uses the recently proposed $\Theta$ criterion to assess local variance contributions \cite{NovVorSadShi:CMAME:21}. The proposed approach balances both \emph{exploitation} of the surrogate model and \emph{exploration} of the input random space and thus leads to efficient and accurate approximation of the original mathematical model. The numerical results show the superiority of the \DALPCE{} in comparison to (i) a~single global polynomial chaos expansion and (ii) the recently proposed stochastic spectral embedding (SSE) method \cite{SSE:Marelli:21} developed as an accurate surrogate model and which is based on a~similar domain decomposition process. This method represents general framework upon which further extensions and refinements can be based, and which can be combined with any technique for non-intrusive polynomial chaos expansion construction.
\end{abstract}

\begin{highlights}
\item Effective construction of a~general purpose surrogate model based on polynomial chaos expansion.
\item Novel method for sequential decomposition of the input random space and construction of local approximations.
\item Sequential domain decomposition and sample size extension based on an active learning methodology.
\item Active learning is represented by variance-based $\Theta$ criterion developed for polynomial chaos expansion.

\end{highlights}


\end{frontmatter}


\section{Introduction}
\label{}

The Polynomial Chaos Expansion (\PCE), originally proposed by Norbert Wiener \cite{Wiener_PCE} and further investigated in the context of engineering problems by many researchers, e.g.~\cite{BLATMANLARS,Ghanem_spectral}, is a~preferred method for uncertainty quantification (UQ) and surrogate modeling in industrial applications \cite{CHEN20084830,fibPCE} thanks to its efficiency and powerful post-processing. Once a~\PCE\ is available for a~given problem, the constructed explicit function can be exploited to directly estimate important properties of the original problem including its statistical moments, response probability distribution or sensitivity indices (without additional sampling \cite{SUDRET}), which brings significant efficiency for surrogate modeling, sensitivity analysis, uncertainty quantification and reliability analysis \cite{CRESTAUX20091161}. 

The \PCE{}, in its non-intrusive form, offers a~convenient way to perform probabilistic analysis of any black-box model, e.g. finite element models representing complex physical systems in engineering. There are generally two types of non-intrusive methods to calculate the deterministic \PCE{} coefficients: spectral projection and linear regression. The spectral projection approach utilizes the orthogonality of the multivariate polynomials and calculates the coefficients using inner products. The spectral projection leads to an explosion of computational complexity referred to as the \emph{curse of dimensionality}. Therefore, the non-intrusive approach based on linear regression is often preferred. Although it is typically less expensive than the spectral projection (the number of samples should be at least  $\mathcal{O}( P \, \ln (P))$, where $P$ is the number of terms in the \PCE{} \cite{CohenOptimalWLS,NarayanOptimalWLS}), it suffers from the \emph{curse of dimensionality} as well, since the number of \PCE{} terms grows rapidly with both dimension and maximum polynomial order. Therefore, it becomes necessary to employ advanced adaptive techniques to construct sparse \PCE{}s that yield efficient solutions for real-world physical systems. 

Regression-based \PCE{} can be significantly affected by the selected sampling scheme, as was recently shown in an extensive review paper \cite{LuthenReview} comparing several general statistical sampling techniques. However, \PCE{} construction as a~linear regression model is a~very problem specific task and it can be highly beneficial to use methods that exploit information from the given mathematical model and sequentially update the surrogate model -- referred to as \emph{active learning}.
Active learning is a~common approach for surrogate-based reliability analysis, wherein an initial experimental design is iteratively updated based on the current estimate of the limit-state surface \cite{Echard2011,Shi2019,Yang2020}. Active learning for reliability analysis with \PCE{} was used e.g. in  \cite{Marelli2018,Zhou2020,Cheng2020}. For general UQ studies, some recent studies have focused on general sequential sampling for \PCE\ based on space-filling criteria or alphabetical optimality \cite{PCESoptimSeq,SequentialPCEThapa}. However, it is beneficial to use both \emph{exploitation} (leveraging model behavior) criteria and \emph{exploration} (space filling) criteria to define an optimally balanced criterion \cite{SHIELDS2018207}. Such sequential sampling for sparse Bayesian learning \PCE{} combining both aspects -- epistemic uncertainty of the statistical inference (exploration) together with quadratic loss function (local exploitation) -- was recently proposed in \cite{ZhouSequential}. However, its application is limited  to \PCE{} built by sparse Bayesian learning only.

The authors of this paper recently proposed a~general active learning method based on sequential adaptive variance-based sampling \cite{NovVorSadShi:CMAME:21}, which is an efficient tool for accurate surrogate modeling that is sufficiently general for further extension \cite{ZHANG2022108749}.
Although this approach leads to superior results in comparison to standard approaches without active learning, it is limited by the inherently smooth nature of the \PCE{}. More specifically, polynomial basis functions are not able to approximate functions with discontinuities or singularities. Moreover, it is necessary to use high-order polynomials to approximate functions with local non-linearities, even when the rest of the input random space could be easily approximated by a~low-order \PCE{}. This can lead to spurious oscillations in the approximation and over-fitting.  
To overcome this limitation, we propose a~method to construct localized \PCE{}s based on the concept of \emph{divide-and-conquer}, i.e. decomposition of the input random space to sub-domains approximated by many low-order \PCE s instead of a~single high-order global \PCE. 
Although this concept is not entirely new in stochastic finite elements \cite{DEB20016359} and stochastic collocation \cite{witteveen2012simplex, BHADURI2018732}, there is no such approach for non-intrusive \PCE. However there are two primary techniques based on similar concepts as described in the following section.

\subsection{Related Developments}

Stochastic Spectral Embedding (SSE) \cite{SSE:Marelli:21} is a~general approximation technique based on a~decomposition of the input random space and the construction of embedded local approximations.
Although it is generally possible to use any spectral approximation technique, it is beneficially coupled with \PCE{}. SSE is based on a~novel idea of \emph{embedding} -- instead of constructing local approximations of the original mathematical model, local surrogates are constructed to approximate the \emph{residuals} between the model and approximation from the previous level of the decomposed space. 
Although such an approach can lead to significant improvement in comparison to a~single global approximation \cite{SSE:Marelli:21}, it is not a~sequential approach based on active learning and thus it does not iteratively reflect new information obtained from the previous steps of the algorithm. Active learning is crucial in analysis of functions with discontinuity or singularity because it allows for the aforementioned exploration and exploitation necessary to find and resolve these features. For the sake of completeness, active learning for SSE has been proposed for reliability analysis \cite{Rare:SSE:WAGNER:22}, but it does not lead to an accurate approximation over the entire input random space. Its accuracy is limited to regions around the limit surface, which are important for an estimation of failure probability.

The second related technique is Multi-element generalized Polynomial Chaos Expansion (ME-gPC) \cite{WAN2005617}. ME-gPC was developed as an extension of generalized \PCE\ based on Wiener-Askey scheme \cite{Askey} allowing analysis of models with arbitrary distribution of input random vector. The ME-gPC method consists of three main parts: decomposition of the input random space, numerical construction of locally orthogonal polynomials and an adaptive procedure based on the decay rate of local error in estimated variance derived from local PCE. ME-PCE applies an $h$-type mesh refinement procedure akin to mesh refinement in finite element methods. By doing so, they introduce a~structured grid of uniform points in each new element and solve for the \PCE{} coefficients. This can be cumbersome and does not afford the flexibility to adaptively select sparse and near-optimal training points.  Moreover, we note that the ME-gPC was created mainly for uncertainty propagation in models with arbitrary input distributions, and thus in contrast to SSE, its objective is not necessarily to construct the best possible surrogate model using adaptive algorithms, but rather to minimize errors in response statistics. This is a~subtle, but important difference that distinguishes its use as a~predictive tool from that of a~tool for statistical estimation.

\subsection{Contributions of this paper}

This paper describes a~novel method, termed Domain Adaptive Localized \PCE{} (\DALPCE) that applies adaptive sequential decomposition of the input random space and adaptive sequential sampling within the sub-domains. Both of these features are based on recently a~proposed criterion for variance-based sequential statistical sampling, developed specifically for \PCE{} in \cite{NOVAK2022106808}. In the context of previously described methods SSE and ME-gPC, the proposed novel approach can be though to lie between them. Like SSE, it is developed specifically for the construction of accurate surrogate models, especially for functions with high non-linearity or discontinuity. But the decomposition of the input random space is rather similar to ME-gPC. The uniqueness of our proposal lies in the combination of active learning, sequential sampling, sequential decomposition of the input space and regression-based \PCE\ using sparse solvers such as Least Angle Regression (LARS) allowing adaptivity and learning in each iteration of the proposed algorithm.

\section{Polynomial Chaos Expansion}
\label{PCE section}

Assume a~probability space ($ \Omega, \pazocal{F}, \pazocal{P} $), where $ \Omega $ is an event space, $ \pazocal{F} $ is a~$ \sigma $-algebra on $ \Omega $ and $ \pazocal{P} $ is a~probability measure on $ \pazocal{F} $.
If the input variable of a~mathematical model, $Y=f(X)$, is a~random variable $ X(\omega) , \omega\in\Omega$,
the model response $Y$($ \omega $) is also a~random variable. Assuming that $ Y $ has a~finite variance, \PCE\ represents the output variable $ Y $ as a~function of an another random variable $ \xi $ called the \emph{germ} with a~known distribution
\begin{equation}
    Y=f(X)=f^{\PCE}(\xi ),
\end{equation}
and represents the function $f(X)$ via infinite polynomial expansion. A~set of polynomials, orthogonal with respect to the distribution of the germ, are used as a~basis of the Hilbert space  $ L^2 $ ($ \Omega,  \pazocal{F}, \pazocal{P} $) of all real-valued random variables of finite variance, where  $ \pazocal{P} $ takes over the meaning of the probability distribution. The orthogonality condition is given by the inner product of $ L^2 $~($ \Omega,  \pazocal{F}, \pazocal{P} $) defined for any two functions $ \psi_j $ and $ \psi_k $ for all  $ j \ne k $ with respect to the weight function $ p_\xi $ (probability density function of $ \xi $) as:
\begin{equation}
    \langle	\psi_j,\psi_k\rangle
    =
    \int \psi_j(\xi)\psi_k(\xi)p_\xi  (\xi)
    \; \mathrm{d} \xi
    = 0.
\end{equation}

This means that there are specific orthogonal polynomials associated with the corresponding distribution of the germ via its weighting function.
For example, Hermite polynomials orthogonal to the Gaussian measure are associated with normally distributed germs.
Orthogonal polynomials corresponding to other distributions can be chosen according to Wiener-Askey scheme~\cite{Askey} or constructed numerically \cite{Gautschi:1982}. For further processing, it is beneficial to use normalized polynomials (orthonormal), where the inner product of $i$th and $j$th polynomials is equal to the Kronecker delta  $ \delta_{jk}$, i.e. $ \delta_{jk}=1$ if and only if $ j=k $, and  $ \delta_{jk}=0 $ otherwise.

In the case of $ \X $ and $\xxi$ being vectors containing $ M $ independent random variables, the polynomial $ \Psi (\xxi)$ is multivariate and it is built up as a~tensor product of univariate orthonormal polynomials, i.e.
\begin{equation}
\label{Eq: MultVarPol}
    \Psi_{\balpha} ( \xxi )
    =
    \prod_{i=1}^{M}  \psi_{\alpha_i}(\xi_i),
\end{equation}
where $ {\balpha}\in \mathbb{N}^M $ is a~set of integers called the \emph{multi-index} reflecting polynomial degrees associated to each $\xi_i$. The quantity of interest (QoI), i.e. the response of the mathematical model $ Y=f(\X)$, can then be represented as~\cite{Ghanem_spectral}
\begin{equation}
\label{PCE}
    Y = f(\X) =
    \sum_{\balpha \in \mathbb{N}^M }
    \beta_{\balpha}\Psi_{\balpha}( \xxi),
\end{equation}
where $ \beta_{\balpha} $  are deterministic coefficients and $ \Psi_{\balpha} $ are multivariate orthonormal polynomials.

\subsection{Non-intrusive computation of \PCE\ coefficients}
For practical computation, the \PCE\ expressed in Eq.~\eqref{PCE} must be truncated to a~finite number of terms $ P $. One can generally choose any truncation rule (e.g. tensor product of polynomials up to the selected order $p$), but the most common truncation is achieved by retaining only terms whose total degree $ \vert \balpha \vert $ is less than or  equal to a~given $ p $, in which case the truncated set of \PCE\ terms is then defined as
\begin{equation}
    \pazocal A^{M,p}
    =
    \left\{
        {\balpha} \in \mathbb{N}^{M} : \left| {\balpha} \right|= \sum_{i=1}^{M} \alpha_i \leq p
    \right\}.
\label{Eq: truncation}
\end{equation}
The cardinality of the truncated \emph{index set} $ \pazocal A^{M,p} $ is given by
\begin{equation}
 \mathrm{card} \: \pazocal A^{M,p}= \frac{\left( M+p \right)!}{M! \: p!}\equiv P \, .
 \label{Eq.: Cardinality PCE}
\end{equation}
When the \PCE\ is truncated to a~finite number of terms, there is an error $ \varepsilon $ in the approximation such that
\begin{equation*}
    Y
    =
    \displaystyle f{(\X)} =
    \sum_{\balpha \in \pazocal A}
    \beta_{\balpha} \Psi_{\balpha}(\xxi)
    +
    \varepsilon  \, .
    \label{Eq: Chaos_Truncated_OLS}
\end{equation*}
From a~statistical point of view, \PCE\ is a~simple linear regression model with intercept. Therefore, it is possible to use \emph{ordinary least squares} (OLS) regression to minimize the error  $ \varepsilon $.

Knowledge of vector $\bbeta$ fully characterizes the approximation via \PCE. To solve for $\bbeta$, first it is necessary to create $ \Ns $ realizations of the input random vector $ \X $ and the corresponding results of the original mathematical model $ \pazocal Y  $, together called the experimental design (ED). Then, the vector of $P$ deterministic coefficients $\bbeta $ can be determined by OLS as
\begin{equation}
    \bbeta
    =
    (\bPsi^{T}\bPsi)^{-1} \ \bPsi^{T}  \pazocal Y ,
\label{Eq: PCe_OLS}
\end{equation}
where $ \bPsi $ is the data matrix
\begin{equation}
    \bPsi
    =
    \left\{
        \Psi_{ij}= \Psi_{j}(\xxi^{(i)}),  \;
        i=1, \ldots, \Ns,   \;
        j=0, \ldots, P-1
    \right\}.
\label{Eq: basis_matrix}
\end{equation}
A well-known problem, the \emph{curse of dimensionality}, states that $ P $ is highly dependent on the number of input random variables $ M $ and the maximum total degree of polynomials $ p $, which is clear from Eq.~\eqref{Eq.: Cardinality PCE}. Considering that estimation of $\bbeta$ by regression requires at least $\mathcal{O}( P \, \ln (P))$ number of samples for stable solution \cite{CohenOptimalWLS, NarayanOptimalWLS}, the problem can become computationally highly demanding in case of a~large or strongly non-linear stochastic models. Although one can use advanced model selection algorithms such as Least Angle Regression (LAR) \cite{LARS,BLATMANLARS}, orthogonal matching pursuit \cite{OMP} or Bayesian compressive sensing \cite{BCS} to find an optimal set of \PCE\ terms, and thus reduce the number of samples needed to compute the unknown coefficients, the benefit of these techniques is significant only if the true coefficient vector is sparse or compressible. The sparse set of basis functions obtained by any adaptive algorithm is further denoted by $ \A $ for the sake of clarity.

\subsection{Approximation Error Estimation}
Once the \PCE\ is constructed, it is crucial to estimate its accuracy. Further, the \PCE\ accuracy can be used to directly compare several \PCE{}s to choose the best surrogate model. Ideally the ED should be divided into validation and training sets, but this might be extremely computationally demanding in engineering applications with complex numerical models. Therefore in the field of uncertainty quantification (UQ) of engineering models, it is preferred to estimate the approximation error directly from the training set, without any additional sampling of the original model. A~common choice is the coefficient of determination $ R^2 $, which is well-known from machine learning or statistics. However, $ R^2 $ may lead to over-fitting and thus advanced methods should be used.  One of the most widely-used methods is the leave-one-out cross-validation (LOO-CV) error $ Q^2 $. The LOO-CV is based on residuals between the original surrogate model and the surrogate model built with the ED while excluding one realization.
This approach is repeated for all realizations in the ED and the average error is estimated. Although the calculation of $ Q^2 $ is typically highly time-consuming, it is possible to obtain results analytically from a~single \PCE\ as follows~\cite{BLATMAN2010}:
\begin{equation}
    Q^2
    =
     \frac{\displaystyle{\frac{1}{\Ns}\sum_{i=1}^{\Ns}
    \displaystyle
    {{{\left[ {\frac{
                g       \left(\xx^{(i)}\right) -
                g^{\PCE}\left(\xx^{(i)}\right)
                }
    {1 - {h_i}}} \right]}^2}} }}
    {\sigma^2 _{Y, \mathrm{ED}}},
    \label{Eq: QQ}
\end{equation}
where ${\sigma^2 _{Y, \mathrm{ED}}} $ is the variance of the ED calculated using the original mathematical model and  ${h_i}  $ represents the $ i $th diagonal term of matrix
$  \mathbf{H}
    =
    \bPsi  \left(
    \bPsi^T
    \bPsi  \right)^{-1}
    \bPsi^T
$.

\subsection{Statistical Moments Derived from \PCE}
\label{moments}
The form of \PCE\ as a~linear summation over orthonormal polynomials allows for powerful and efficient post-processing. In particular, once a~\PCE\ approximation is created, it is possible to directly estimate statistical moments of the output from the expansion.

The first statistical moment (the mean value) is simply the first deterministic coefficient of the expansion $\mu_{Y}=\big<Y^{1} \big> = \beta_{\bzero}$. The second raw statistical moment, $ \big<Y^{2} \big> $, can be estimated by
\begin{align}
    \label{Eq:SecondRawMoment}
    \left\langle {{Y^2}} \right\rangle
    &=
    \int
    {\left[ {\sum\limits_{\balpha \in \A}
        {{\beta _{\balpha}}{\Psi _{\balpha}}\left( \xxi  \right)} } \right]} ^2
        p_{\xxi} \left( \xxi \right)   \;
        \mathrm{d}  \xxi =
    \sum\limits_{\balpha_1 \in \A}
    \sum\limits_{\balpha_2 \in \A}
    \beta_{{\balpha}_1}
    \beta_{{\balpha}_2}
    \int
    \Psi_{{{\balpha}_1}}\left( \xxi \right)
    \Psi_{{{\balpha}_2}}\left( \xxi \right)
    p_\xxi \left( \xxi \right)  \;
    \mathrm{d} \xxi
    \\   \nonumber
    & =
    \sum\limits_{{\balpha} \in {\A}} {\beta_{\balpha}^2} {\int {{\Psi _{\balpha}}\left( \xxi \right)} ^2}{p_\xxi}
    \left( \xxi\right)  \;
    \mathrm{d}  \xxi =
    \sum\limits_{{\balpha} \in {\A}} {\beta_{\balpha}^2} \left\langle {{\Psi _{\balpha}},{\Psi _{\balpha}}} \right\rangle.
\end{align}
Considering the orthonormality of the polynomials, it is possible to obtain the variance  $ 	\sigma_{Y}^{2}=\big<Y^{2} \big>- \mu_{Y}^{2} $
 as the sum of all squared deterministic coefficients except the intercept (which represents the mean value), i.e.
\begin{equation}
    \label{Eq: Variance}
    \sigma_{Y}^{2}
    =\sum_{\substack{\balpha \in \A\\ \balpha\neq \bzero }  }
    \beta_{\balpha}^{2}.
\end{equation}
Note that the computation of higher statistical central moments, specifically 
skewness $\gamma_Y$ ($ 3^{\text{rd}}$ moment) and kurtosis $\kappa_Y$ ($ 4^{\text{th}}$ moment), are more complicated since they require triple and quad products. These can be obtained analytically only for certain polynomial families, e.g. formulas for Hermite and Legendre polynomials (and their combination) can be found in \cite{NOVAK2022106808}.

\section{Active Learning-based Domain Adaptive Localized PCE (\DALPCE)}

In this section, we propose a~novel methodology to constructed localized \PCE{}s designed for highly non-linear functions, termed Domain Adaptive Localized \PCE{} (\DALPCE). Instead of increasing the maximum polynomial order $p$ ($p$-adaptivity), which brings high computational requirements due to the \emph{curse of dimensionality}, we propose to decompose the input random space into several sub-domains approximated by low-order \PCE s ($h$-adaptivity). Although this idea is not entirely new, we use this approach in combination with novel active learning methods to identify domains for refinement and for sequential sample selection and regression-based \PCE s. This allows us to use any sparse adaptive solver (e.g. LAR) and thus it can be easily implemented into the existing software packages \cite{olivier2020uqpy,UQlab}. In the following sections, we define the requisite components of the proposed method and provide an algorithm (Algorithm \ref{alg:1}) for its implementation.

\subsection{Variance-based Adaptive Sequential Sampling}
The decomposition of the input random space is a~sequential process coupled with adaptive sampling assuring optimal coverage of the sub-domains of interest. The whole process thus consists of two steps: (i) identification of an important sub-domain, that is, a~domain that is either large compared to other sub-domains or that is  associated with a~high local variance; and (ii) identification of the best positions for additional samples extending the current ED in the selected sub-domain. Each of these steps must be based on a~criterion that balances \emph{exploration} of the input random space with \emph{exploitation} of the surrogate model, which in our case is in the form of a~\PCE.  The $\Theta$-criterion for adaptive sequential sampling, which is driven by the output variance and its approximation via local variance using \PCE\ \cite{NovVorSadShi:CMAME:21}, is employed for both steps. We will first discuss the process for adaptive sequential sampling within a~specified sub-domain in this section. This will be followed by the process for refinement of the domain in the subsequent sections. 

Consider a~pool of candidate samples containing realizations of the random vector $ \xxi$ generated by an arbitrary sampling technique, e.g., Latin Hypercube Sampling (LHS) \cite{McKayConovBeck:three:1979,Conover:LHS:75} or Coherence sampling \cite{CoherenceOptPCE,CohDOpt,CohenOptimalWLS}. 
From this pool of candidates, we select the best sample using a~method inspired by the sequential sampling proposed in \cite{SHIELDS2018207} and based on Koksma-Hlawka inequality \cite{Koksma:ineq_1942}. The $\Theta$-criterion for \PCE, which accounts for both variation of the function and discrepancy of the samples, was proposed as follows \cite{NovVorSadShi:CMAME:21}:
\begin{align}
    \label{Eq:ThetaCrit}
    \Theta (\xxi^{(\cand)}) \equiv \Theta^c
    &=
    \underbracket[0.4pt]{
                \sqrt{\sigma_{\!\!\A}^{2}(\xxi^{(\cand)})\cdot\sigma_{\!\!\A} ^{2}(\xxi^{(\neigh)})}
                }
                _{\mathrm{ave \; variance \; density}}
    \:
     \underbracket[0.4pt]{
                l_{\cand,\neigh}^M
                }_{\mathrm{vol.}}
\equiv
                \sqrt{\sigma_{\cand}^{2} \cdot \sigma_{\neigh}^{2}}
    \:
    l_{\cand,\neigh}^M.
\end{align}
The criterion is a~product of two terms -- the \emph{exploitation} term (denoted as ``ave variance density'') and the \emph{exploration} part (the distance term $l_{\cand,\neigh}$ raised to the domain dimension) -- which are multiplied to maintain an optimal balance between exploration and exploitation  \cite{NovVorSadShi:CMAME:21}.

The \emph{exploration} aspect is maintained by accounting for the distance $ l_{\cand,\neigh}$ between a~candidate $ \xxi^{(\cand)} $ and its nearest neighboring realization from the existing ED, $ \xxi^{(\neigh)} $ as
\begin{equation}
    \label{eq:metric}
    l_{\cand,\neigh}=\sqrt{\sum_{i=1}^{M} |\xi_i^{(\cand)}-\xi_i^{(\neigh)}|^2  }.
\end{equation}
If the criterion was reduced to this term only, sequential filling of the greatest empty regions would occur, converging to uniform space coverage in the spirit of the space-filling ``miniMax criterion'' \cite{JohMooYlv:MixiMinMiniMax:JSPI:90,Pronzato:MinimAndMaxim:17,EliVorSad:miniMax:ADES:20}.

The \emph{exploitation} component is motivated by the desire to sample points in regions with the greatest contributions to the total variance of the QoI $\sigma_{Y}^{2}$, i.e. at points with the highest \emph{variance density}. Once the \PCE\ has been established at any given stage of the algorithm, the \emph{variance density} is computationally cheap to evaluate for any location $\xxi$ as
\begin{equation}
    \sigma_{\!\!\A} ^{2}(\xxi)
    =\big[ \sum_{\substack{\balpha \in \A \\ \balpha\neq\bzero}  }
        \beta_{\balpha}
       {\Psi_{\balpha}
        \left(\xxi\right)} \big]^2
    {p_\xi }\left( \xxi\right).
\end{equation}
The local variance is therefore estimated directly using the basis functions and coefficients $ \beta $ of the \PCE.  When considering a~candidate ``$\cand$'', an estimate of the variance contribution of the region between the candidate and its nearest neighbor ``$\neigh$'' may be obtained by averaging the local variance densities between the two. Therefore, we can say that the candidate with the greatest $\Theta^c$ criterion is the one that represents the largest amount of total variance to be refined by its selection. 

A significant advantage of this method is the ability to add candidates into an existing ED one-by-one. Thus, it can be employed at any moment of the \PCE\ construction process. Moreover, this learning function can be combined with any sampling algorithm for the construction of the initial ED and candidates for extension. The ideas behind the $\Theta$ criterion will now be used in the proposed domain decomposition and ED extension algorithm.

\subsection{Decomposition of Input Random Space}

\begin{figure}[h] 
       \centering
	\includegraphics[width=0.8\textwidth]{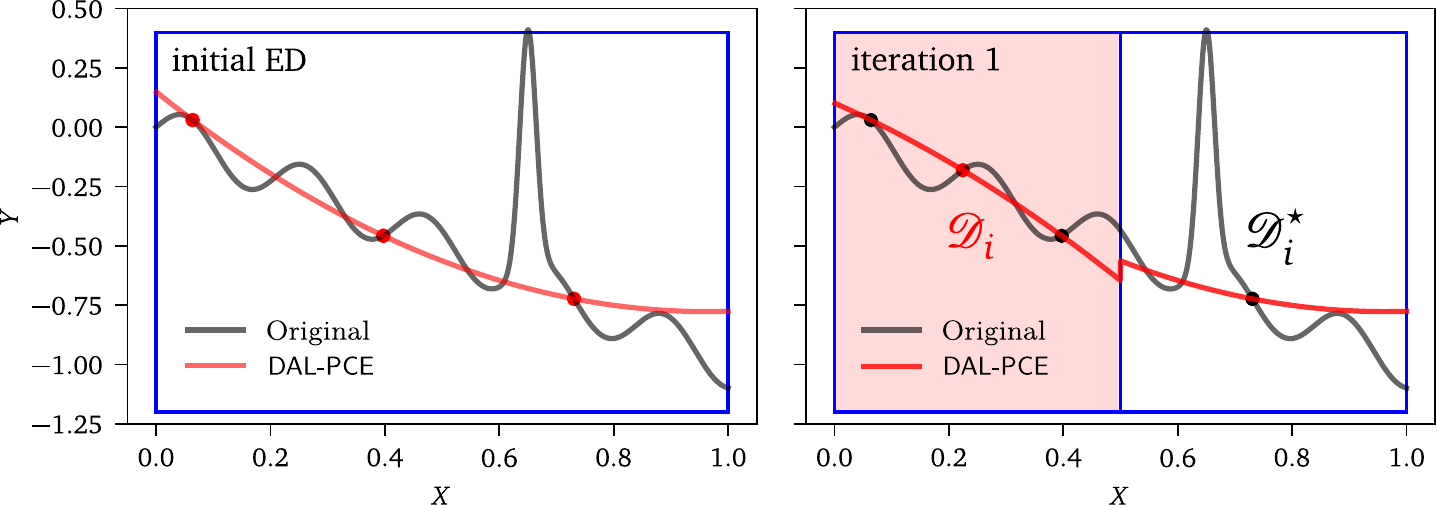}
	\caption{The first iteration of the algorithm: the original sub-domain is split and the new local \PCE\ is constructed in $\mathcal{D}_i$ (red background), while the second part in $\mathcal{D}_i^\star$ inherits the \PCE\ approximation from the original domain.}
	\label{Fig: Algorithm_overlap}
\end{figure}

\begin{figure}[bth] 
    \centering
	\includegraphics[width=1\textwidth]{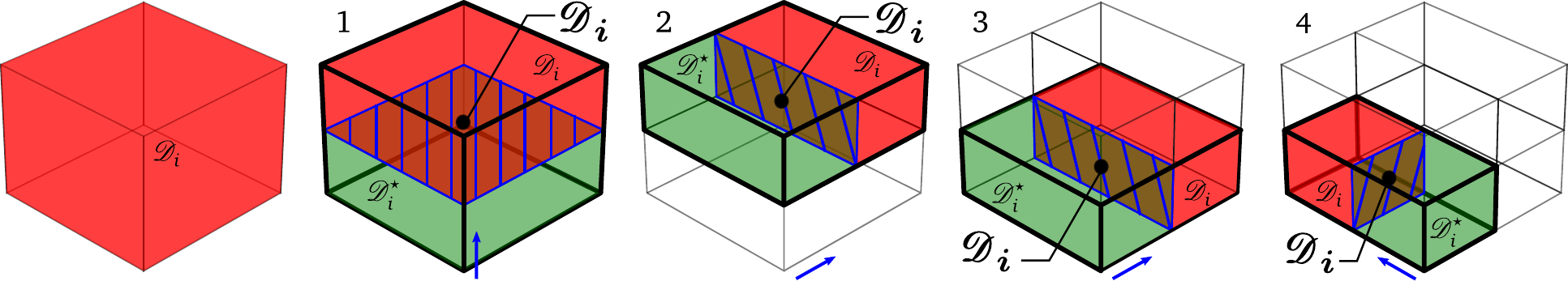}
	\caption{The first four steps of the decomposition of a~3D space of input random variables. The thick black lines outline the parent domain selected for division.
             The red and green boxes inside it represent the two newly created refinement-child $\mathcal{D}_i$ (red) and inheriting-child $\mathcal{D}_i^\star$ (green) sub-domains created by splitting the parent domain $\bm{\mathcal{D}_i}$ (bold boundaries), selected via Eq.~\eqref{Eq:ModifiedThetaCrit}, by the cutting plane (blue). The cutting plane is perpendicular to the variable selected for splitting (blue arrow).
             }
	\label{Fig:Decompositon}
\end{figure}

The core of the proposed approach is a~sequential decomposition of the input random space $\mathcal{D}$ for the construction of local approximations. This approach assumes that the original mathematical model can be approximated by piecewise low-order \PCE s that are valid only in individual sub-domains of $\mathcal{D}$. 
Therefore, in the proposed approach, the input random space is sequentially decomposed into $n_{\mathcal{D}}$ smaller non-overlapping sub-domains $\mathcal{D}_i \subset \mathcal{D}$ that collectively fill the full input random space $\mathcal{D}$, i.e.
\begin{equation}
    \bigcup_{i=1}^{n_{\mathcal{D}}} \mathcal{D}_i = \mathcal{D} \quad \text{such that} \quad \mathcal{D}_i \cap \mathcal{D}_j = \emptyset \quad \forall i,j
    \label{Eq.: Decomposition}
\end{equation}
In each iteration of the algorithm, a~single sub-domain $\bm{\mathcal{D}_i}$ (referred to as the parent) is identified for refinement and divided by a~plane perpendicular to the direction of one selected input random variable. Specifically, $\bm{\mathcal{D}_i}$ is divided into a~refinement-child $\mathcal{D}_i$, which is further processed, and an inheriting-child $\mathcal{D}_i^{\star}$ adopting the \PCE{} from the parent as illustrated for a~one-dimensional function in Fig. \ref{Fig: Algorithm_overlap}. In this case, we see that the space is divided into two subdomains. In the left (refinement child) a~new \PCE{} is constructed. In the right (inheriting child), the original PCE is retained. Such process assures an exhaustive decomposition into disjoint subsets i.e.~$\bm{\mathcal{D}_i}=\mathcal{D}_i\oplus\mathcal{D}_i^{\star}$. 
This sequential domain decomposition is illustrated in Fig.~\ref{Fig:Decompositon}, which depicts the original input random space and the first four iterations of the decomposition process. 

In contrast to SSE \cite{SSE:Marelli:21}, the selection of a~single sub-domain for refinement in each iteration is based on an active learning approach, the details of which are provided in subsequent sections. Importantly, actively integrating information from the original mathematical model leads to a~significantly more effective decomposition of the space and thus assures accurate approximations, even for small-size EDs. On the other hand, the identified decomposition and the associated ED are directly connected to the given mathematical model and therefore might be inefficient for general statistical analysis. 

The complete surrogate model is assembled from the $n_{\mathcal{D}}$ local \PCE s associated with all sub-domains $\mathcal{D}_i$ as:
\begin{equation}
    Y \approx \sum_{i=0}^{n_{\mathcal{D}}}
    \sum_{\balpha_i \in \pazocal A_i}
    \beta_{\balpha_i} \Psi_{\balpha_i}(\xxi) \mathbb{1}_{\mathcal{D}_i}(\xxi),
    \label{Eq: Chaos_Truncated_OLS}
\end{equation}
where $\mathbb{1}_{\mathcal{D}_i}(\xxi)$ represents indicator function, i.e. $\mathbb{1}_{\mathcal{D}_i}(\xxi)=1$ only if $\xxi\in\mathcal{D}_i$ and $\mathbb{1}_{\mathcal{D}_i}(\xxi)=0$ otherwise. In other words, to  approximate the original model at any point, it suffices to determine the one relevant sub-domain and use the corresponding local \PCE. Each such local \PCE\ has its own set of basis functions $\pazocal A_i$ and corresponding coefficients $ \beta_{\balpha_i}$, which can be obtained by any model-selection algorithm. In this paper the OLS and LAR algorithms are employed, but generally any non-intrusive technique can be used.

\subsection{Domain Selection via Modified Variance-based Criterion}
The selection process to identify the ``best'' subdomain for possible division is governed by extending the $\Theta$-criterion from Eq.\ \eqref{Eq:ThetaCrit} as follows:
\begin{equation}
\label{Eq:ModifiedThetaCrit}
    \Theta_{i} = 
    \underbracket[0.4pt]{\pazocal{W}_i\cdot \exp{(Q_{i}^2)} 
                        }_{\mathrm{weight \: of \: subdomain}}
    \cdot
    \underbracket[0.4pt]{
	\sqrt{\sigma_{\!\!\A_i}^{2}(\boldsymbol \xi^{(c)})\cdot\sigma_{\!\!\A_i} ^{2}(\boldsymbol \xi^{(s)})} \: l_{c,s}^M}
        _{
            \Theta^c \: \text{in  \:} i\text{th  subdomain}
        }.
\end{equation}
This extended criterion aims to identify sub-domains of the input random space associated with the maximum value of $\Theta^c$, while simultaneously accounting for the size of each subdomain and the accuracy of the existing local \PCE. The former is calculated using Eq.~\eqref{Eq:ThetaCrit}  calculated for a~rich pool of screening global candidates, while the latter are measured by incorporating the volume of each sub-domain $\pazocal{W}_i$ and the LOO-CV error $Q_{i}^2$, respectively. 
The LOO-CV term, $\exp{(Q_{i}^2)}$, can be thought to artificially inflate the domain volume as a~penalization for inaccurate approximation. When the approximation is perfect $(Q_{i}^2=0)$ the true volume of the sub-domain is used. Meanwhile, a~poor approximation with $Q_{i}^2=1$ leads to roughly 2.72 times increased volume. 

The three terms featured in Eq.~\eqref{Eq:ModifiedThetaCrit} aim at different aspects affecting the accuracy of the final surrogate model: 
    large sub-domains are preferred by $\pazocal{W}_i$, 
    sub-domains containing poor \PCE\ approximation are promoted via $\exp{(Q_{i}^2)}$ and finally, 
    $\Theta^c$ prefers sub-domains with high concentration of variance. 
Note that $\Theta^c$ is calculated for a~rich pool of screening candidates, and  $\pazocal{W}_i$ and $\exp{(Q_{i}^2)}$ are calculated directly from the geometry of existing sub-domain and the local \PCE{} model, respectively.  The product of all three terms in the extended criterion therefore maintains the desired balance and assures the selection of the sub-domain, $\mathcal{D}_i$, that currently seems to be the most important for increasing the accuracy of the \PCE\ surrogate model. 

Sub-domain $\mathcal{D}$ with the greatest $\Theta_{i}$ is selected and one of the operations described in detail in Sec.~\ref{sec:algo} is performed, depending on whether $\mathcal{D}_i$ contains a~critical number of ED points. Two scenarios can occur:
\begin{itemize}
\item
    $\mathcal{D}_i$ contains a~sufficient number of ED points ($n_i \geq \ns$) to ensure accuracy of a~\PCE{} on the domain. Therefore, it becomes a~parent $\bm{\mathcal{D}_i}$ (bold boundaries in Fig.~\ref{Fig:Decompositon}) and is divided into two parts by a~selected rule. The child domain containing the decisive candidate with the greatest $\Theta^c$ becomes the refinement-child $\mathcal{D}_i$ (see the red subdomains in steps $1-4$ in Fig.~\ref{Fig:Decompositon}). 
    The remaining volume becomes an inheriting-child denoted $\mathcal{D}_i^{\star}$ (see the green subdomains in Fig.~\ref{Fig:Decompositon}), which retains the \PCE{} from the parent. Division occurs by a~cutting plane, oriented perpendicular to the selected direction (blue arrows in Fig.~\ref{Fig:Decompositon}) and naturally, the coordinates of the cutting plane are restricted to the bounding box of the selected parent $\bm{\mathcal{D}_i}$, see Sec.~\ref{sec:algo}. If needed, the refinement-child domain $\mathcal{D}_i$ is sequentially filled with additional ED points (according to $\Theta^c$) to reach $n_i = \ns$ needed to construct a~new \PCE\ approximation.
    
\item
    $\mathcal{D}_i$ does \emph{not} contain a~sufficient number of ED points ($n_i <  \ns$). The domain is not divided because the suggestion for division is based on insufficient information. Instead, new ED points are sequentially added to $\mathcal{D}_i$, again using the $\Theta^c$ criterion. Note that this scenario practically arises when the selected domain was an inheriting-child in the previous iteration. In this case, the selected domain has inherited a~\PCE{} model that was constructed over a~larger domain. When that domain was divided, it was left with an insufficient number of points from which to construct a~new \PCE{}.
    
\end{itemize}

\subsection{PCE Basis Functions}

Without loss of generality, the proposed method operates on the $M$-dimensional unit hypercube with uniform distributions of input random variables, i.e. $\X\sim \pazocal{U} [0,1]^M$. In the case of a~general joint probability distribution of $\pmb{X}$, it is always possible to transform input random vector to the unit hypercube by Rosenblatt transformation \cite{ROSENBLATT}, Nataf transformation \cite{Nataf} or various methods based on copulas \cite{CopulaFramework}. Standard normalized Legendre polynomials, orthonormal to the uniform distribution, can thus be used as basis functions for the \PCE{}. However, due to the decomposition of the input random space to smaller sub-domains, each with lower bound $a_i$ and upper bound $b_i$, it is necessary to use univariate scaled orthonormal Legendre polynomials of $n$th order $\tilde{\psi}_n(\xi)$ defined as follows:
 \begin{equation}
 \label{Eq.: ScaledLegendre}
 \tilde{\psi}_n(\xi) = \psi_n\left(\frac{2\xi-a_i-b_i}{b_i-a_i}\right),
 \end{equation}
where $\psi_n$ represents standard orthonormal Legendre polynomials. Naturally, the transformation of the original input random vector to the unit hypercube might bring additional non-linearity, and thus one might prefer the direct construction of polynomials locally orthonormal to the given original probability measure as proposed in the Me-gPC \cite{WAN2005617}. While certainly possible, this brings additional computational demands and thus it is not employed here.

\subsection{Local and Global Statistical Estimates from \DALPCE}

The significant advantage of \PCE\ is that analytically post-processing of the expansion yields highly efficient estimates of statistical moments \cite{NOVAK2022106808}, sensitivity indices \cite{SUDRET} and LOO-CV \cite{BLATMANLARS}. In the proposed \DALPCE, since the original domain $\mathcal{D}$ is decomposed into a~set of sub-domains (see Eq.~\eqref{Eq.: Decomposition}), standard analytical post-processing can be applied locally and global characteristics can be obtained by simple weighted summations that converge to the true values as $n_{\mathcal{D}}$ increases.
Specifically, the global mean value and variance of a~QoI are obtained from localized \PCE s (denoted by subscript $\mathcal{D}_i$) as follows:
\begin{equation}
    \label{Eq: Local Mean}
    \mu_{Y}
    =\sum_{i=1}^{n_{\mathcal{D}}}\pazocal{W}_i \beta_{0_i}=\sum_{i=1}^{n_{\mathcal{D}}}\pazocal{W}_i \mu_{\mathcal{D}_i},
    \
\end{equation}
\begin{equation}
    \label{Eq: Local Variance}
    \sigma_{Y}^{2}
    =\sum_{i=1}^{n_{\mathcal{D}}}\pazocal{W}_i\sum_{\substack{\balpha_i \in \A_i\\ \balpha_i\neq \bzero }  }
    \beta_{\balpha_i}^{2}=\sum_{i=1}^{n_{\mathcal{D}}}\pazocal{W}_i \sigma_{\mathcal{D}_i}^{2}.
\end{equation}
where the local mean $\mu_{\mathcal{D}_i}$ and variance $\sigma_{\mathcal{D}_i}^{2}$ are obtained as described in Section \ref{moments}. 

Local Sobol' indices, $S_{\mathcal{D}_i}$, of any order can be derived directly from localized \PCE s and their first-order (main effect) estimates are given by
\begin{equation}
S_{\mathcal{D}_i}^{X_j} = \frac{1}{{\sigma_{\mathcal{D}_i}^{2}}} {{\sum\limits_{\balpha_i \in \A^{X_j}_i} {\beta _{\bf{\balpha_i }}^2} }}{}\;\quad {\A^{X_j}_i} = \left\{ {{\balpha_i} \in {\A_i}:{\alpha _i^j} > 0,{\alpha_i ^{k \ne j}} = 0} \right\}.
\end{equation}
These local Sobol' indices are used in the \DALPCE{} to determine the cut direction (see Section \ref{sec:algo}). Likewise, global Sobol' indices can be obtained easily from weighted summation of local contributions to partial variances normalized by $\sigma_{Y}^{2}$ as follows:
\begin{equation}
S_{X_j} = \frac{\sum_{i=1}^{n_{\mathcal{D}}}\pazocal{W}_i {{\sum\limits_{\balpha_i \in \A^{X_j}_i} {\beta _{{\balpha_i }}^2} }}}{\sigma_{Y}^{2}}
.
\label{eqn:Local_Sobol}
\end{equation}
Similarly, global LOO-CV, $Q^2$, of a~QoI can be approximated by the weighted summation of the local contributions as 
\begin{equation}
    Q^2
    =\sum_{i=1}^{n_{\mathcal{D}}}\pazocal{W}_i
    Q^2_{\mathcal{D}_i},
    \label{Eq: Local QQ}
\end{equation}
where $Q^2_{\mathcal{D}_i}$ are obtained from each local \PCE{} using Eq.\ \eqref{Eq: QQ}.

These estimates are used throughout the proposed \DALPCE{}, as described in detail next.

\subsection{Numerical Algorithm
\label{sec:algo}
}

Based on the presented theoretical background, we now present the numerical algorithm for the domain adaptive localized \PCE. As mentioned above, the whole process can be divided to two iterative tasks: (i) decomposition of the input random space and (ii) construction of localized \PCE s. Both of these tasks are described in the following paragraphs with specific reference to the steps in Algorithm \ref{alg:1}. 

\begin{algorithm}[!ht]
	\caption{\DALPCE: Active Domain Decomposition and Construction of Localized \PCE s}
 \label{alg:1}
\begin{algorithmic}[1]
\Statex	\textbf{Input:}  maximum local polynomial order $p$, number of screening global candidates $n_{c,g}$, number of local candidates $n_{c,l}$, number of iterations $n_\mathrm{iter}$
\State set the minimum number of realizations for local \PCE\ construction $\ns \in \langle P, 2P \rangle$
\State generate a~rich pool of $n_{c,g}$ screening candidates
\State generate the initial ED (size $\ns$) and construct the initial global \PCE

\For{$1$ to $n_\mathrm{iter}$}
\State	identify the sub-domain $\mathcal{D}_i$ with the highest $\Theta_{i}$ based on screening candidates 
\State $n_i \leftarrow$ number of ED samples existing in $\mathcal{D}_i$
\If{$n_i\geq \ns$}
    \State the identified sub-domain $\mathcal{D}_i$ becomes a~parent $\bm{\mathcal{D}_i}$
    \State identify the direction of the highest first-order Sobol' index $S_{\mathcal{D}_i}$ of the parent $\bm{\mathcal{D}_i}$
    \State restrict coordinates of $\bm{\mathcal{D}_i}\rightarrow\mathcal{D}_i$ and create $\mathcal{D}_i^\star$ 
    \State $n_i \leftarrow$ number of ED samples existing in $\mathcal{D}_i$
\EndIf
\State generate $n_{c,l}$ local candidates in $\mathcal{D}_i$
\While{$n_i<\ns$ }
    \State extend size of local ED $n_i$ using the local $\Theta^c$ criterion
\EndWhile
\State  reconstruct local \PCE s in the $\mathcal{D}_i$
\EndFor

\Statex	\textbf{Output:} list of subdomains and corresponding \PCE s

\end{algorithmic}
\label{alg:1}
\end{algorithm}

The first task identifies the important sub-domain $\mathcal{D}_i$ that should be divided and over which low-order local \PCE\ should be constructed. The sub-domain $\mathcal{D}_i$ is specifically identified using the $\Theta_i$ criterion from Eq.~\eqref{Eq:ModifiedThetaCrit}, which again incorporates three important characteristics for accurate surrogate modeling -- 
   the size of the sub-domain $\pazocal{W}_i$, 
   the accuracy of the existing local \PCE\ measured by $Q^2_{\mathcal{D}_i}$, and the original $\Theta^c$ criterion measuring the variance contribution in $\mathcal{D}_i$. 
While $\pazocal{W}_i$ and $Q^2_{\mathcal{D}_i}$ are computed for the whole sub-domain, $\Theta^c$ is computed at specific realizations of input random vector. Therefore, it is necessary to  cover the sub-domains by a~sufficiently large number of screening candidates, such that the total global number of screening candidates is given by $n_{c,g}$. Based on numerical experiments, we recommend $n_{c,g}\ge1000\,M$ to ensure that each sub-domain contains a~sufficient number of screening candidates. Note that the screening candidates are used only to identify $\mathcal{D}_i$ [\emph{step~5}]. They are not used for the ED, and thus even high $n_{c,g}$ does not bring any additional computational demand. 

Once $\mathcal{D}_i$ is identified, it is necessary to check whether there are enough samples to construct a~\PCE\ inside the sub-domain. We start with finding out how many points belong to the selected domain $\mathcal{D}_i$ [\emph{step 6}]. If the number of samples in the identified sub-domain, $n_i$, is greater than  (or equal to) $\ns$ [\emph{step 7}], a~local \PCE\ already exists for $\mathcal{D}_i$. The subdomain is then assigned as a~parent $\bm{\mathcal{D}_i}$ for division [\emph{step 8}] and the first-order Sobol' indices are estimated by Eq.\ \eqref{eqn:Local_Sobol} [\emph{step 9}].  This identified parent $\bm{\mathcal{D}_i}$ is divided in the direction of the highest first-order Sobol' index $S_{\mathcal{D}_i}^{X_j}$. The new restricted coordinates of refinement-child $\mathcal{D}_i$ are identified and the inheriting-child $\mathcal{D}_i^\star$ is created [\emph{step 10}]. Further, the number of ED samples  $n_i$ in the refinement-child $\mathcal{D}_i$ is determined [\emph{step 11}]. On the other hand, if the identified sub-domain $\mathcal{D}_i$ does not contain enough samples (i.e. $n_i< \ns$), the inherited \PCE{} from the previous iteration is not sufficiently local (it was trained over a~domain that has since been divided) and it is necessary to add new samples to $\mathcal{D}_i$ before constructing a~new local \PCE.

The second task of the proposed algorithm is sequential sampling and adaptive \PCE\ construction in sub-domain $\mathcal{D}_i$. Recall that this domain may be either 
\begin{enumerate}[label=(\roman*)]
    \item a~refinement-child that was just divided but does not contain a~sufficient number of points ($n_i < \ns$) or, 
    \item an inheriting-child that now does not contain at least $\ns$ ED samples.
\end{enumerate}
Next, a~set of local candidates is generated in region $\mathcal{D}_i$ [\emph{step 13}]. To ensure sufficient assessment of the coverage of the domain, the number of local candidates is empirically recommended as $n_{c,l} \in \langle 3P,5P \rangle$ \cite{NovVorSadShi:CMAME:21}. From these candidates, the standard $\Theta^c$ criterion in Eq.~\eqref{Eq:ThetaCrit} is used to iteratively select the best candidates until there are $\ns$ samples in $\mathcal{D}_i$ [\emph{step 14-16}]. 
This sequential extension of the sample in $\mathcal{D}_i$ is adaptive in the sense that the pairwise distances in Eq.~\eqref{Eq:ThetaCrit} between candidates and existing ED points are updated after the addition of each new point. However, because $n_i< \ns$ the local variance densities are estimated from the previously existing \PCE, which cannot be updated until a~sufficient number of samples are available in $\mathcal{D}_i$.

The last step of each iteration is to construct the local \PCE\ using scaled Legendre polynomials as basis functions (see Eq.~\eqref{Eq.: ScaledLegendre}) [\emph{step 17}]. Any non-intrusive technique can be used to estimate the coefficients $\bbeta$; we use LARS and OLS for an adaptive construction of the local \PCE s in this paper. At the end of the iteration, all sub-domains are re-numbered and a~list of sub-domains with corresponding \PCE s can be exported or the next iteration can be started.

\subsection{Adaptivity in \PCE\ Construction and Domain Decomposition}

Adaptivity is central to the proposed \DALPCE{}. In the proposed algorithm, there are two types of adaptivity employed: 
\begin{enumerate}[label=(\roman*)]
    \item adaptivity in \PCE\ construction (selection of the optimal set of basis functions), and
    \item adaptivity in domain decomposition
\end{enumerate}
Since the \PCE\ can be constructed by any regression technique in each sub-domain, \PCE\ adaptivity is incorporated by sparse solvers and best model selection algorithms, e.g. Least Angle Regression \cite{LARS}, orthogonal matching pursuit \cite{OMP} or Bayesian compressive sensing \cite{BCS}. Although sparse solvers are often used for \PCE\ with high~$p$, this adaptivity is also important for reducing the number of basis functions (and thus the minimum number of ED samples) for high-dimensional examples or, in our case, for very low-size ED in each $\mathcal{D}_i$ approximated by low-$p$ local \PCE.

\begin{figure}[h] 
\centering
	\includegraphics[width=1\textwidth]{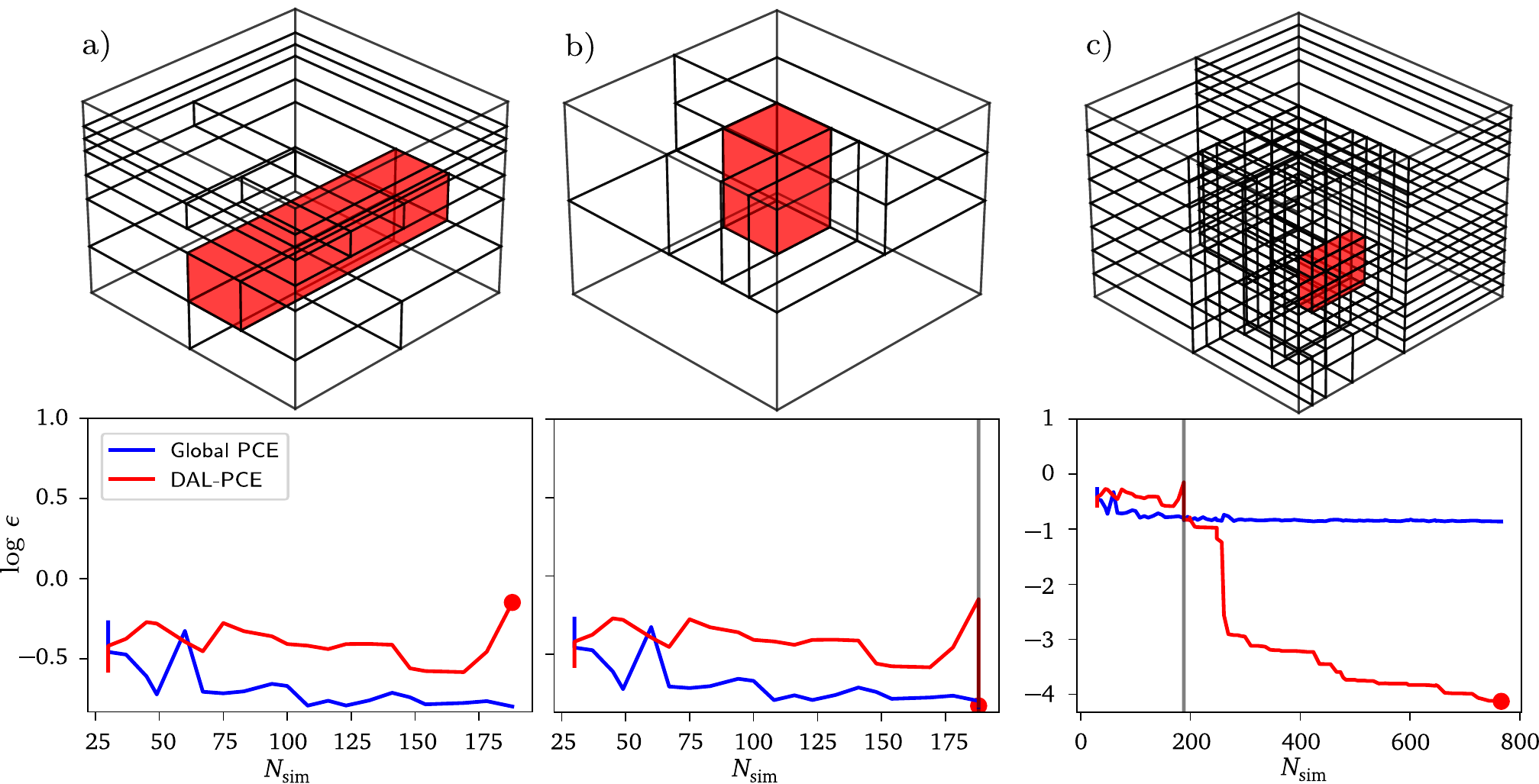}
	\caption{Illustration of domain decomposition restart. 
    a) decomposition and error evolution prior to restart, 
    b) rebuilt decomposition and error drop right after the restart, 
    c) final decomposition and error showing that the restart unlocks a~dramatic decrease in approximation error.
 }
	\label{Fig: restart_decompostion}
\end{figure}

The second type of adaptivity is the proposed adaptivity in the domain decomposition. At any point in the iterative process, the existing ED samples can be used to construct local \PCE{}s or
a~single global \PCE. The \DALPCE{} is not guaranteed to provide a~better approximation than the global \PCE{}. 
This can be measured via $Q^2$, specifically by computing $Q^2_\mathrm{local}$ from Eq.~\eqref{Eq: Local QQ} and $Q^2_\mathrm{global}$ from a~single global \PCE\ according to Eq.~\eqref{Eq: QQ}. If $Q^2_\mathrm{local}>Q^2_\mathrm{global}$ at a~given iteration, the domain decomposition is deemed to be poor and the whole decomposition process is \emph{re-started}. That is, the complete geometrical decomposition is forgotten and all existing ED points are taken as an initial ED for a~brand new run of the algorithm. 
This is illustrated in Fig.~\ref{Fig: restart_decompostion} which shows 
        the decomposition (top) and the associated error (bottom) right before the restart a) at $\Ns = 181$, 
        b) the new decomposition and error right after the restart, 
        and c) the final decomposition/error which shows significant improvement over the global \PCE{}. 
These histories show the standard $R^2$ error defined in Eq.~\eqref{Eq: Error measure}. 
It is not necessary to check this criterion at every iteration, but it is suggested to check it periodically, every $n_{r}$ steps, to ensure adequate local refinement.    

\subsection{Stopping Criteria}
The proposed \DALPCE{} algorithm can be fully automated by adding an adequate stopping criterion. A~simple but practical stopping criterion is based on computational budget, i.e. once the total number of model evaluations $\Ns$ or number of iterations $n_\mathrm{iter}$ have reached a~critical level/budget. 
One may also use a~stopping criterion based on decomposition pattern, e.g. the smallest or the largest volumes of any subdomain, to ensure a~desired resolution. Valuable stopping criterion can be also obtained directly from $Q^2$, corresponding to a~target/threshold level of achieved accuracy. Regardless of the selected stopping criteria, it can easily be applied before \textit{step} 5 of the proposed algorithm (start of each iteration).

\section{Numerical Experiments}

The proposed \DALPCE{} is presented on four numerical examples of increasing complexity and which illustrated different aspects of the approach. The obtained results are compared 
    (a) to the standard global \PCE\ approach with adaptive maximum order $p\in[5,25]$ and 
    (b) to SSE \cite{SSE:Marelli:21}, as current state-of-the-art  non-intrusive surrogate modeling technique based on the domain decomposition. 
The \PCE\ is constructed using the \textsf{UQPy} package \cite{olivier2020uqpy} and the original implementation of SSE is used from the \textsf{UQLab} package \cite{UQlab}. To compare methods, the relative mean squared errors $\epsilon$ are calculated for all three approximations $\tilde{f}$ on a~validation set containing a~large pool of $10^6$ integration points generated by crude Monte Carlo according to:
\begin{equation}
  \epsilon(\X) \coloneqq \frac{\mathbb{E}\Big[\big(f(\X)-\tilde{f}(\X)\big)^2\Big]}
                                 {\mathbb{D}\Big[f(\X)\Big]},
  \label{Eq: Error measure}
\end{equation}
where $\mathbb{E}[]$ and $\mathbb{D}[]$ are the mean value and variance operators, respectively.

To show representative results of the proposed \DALPCE{} algorithm, the calculations were repeated 100 times, and the same settings of the algorithm for all examples were selected as follows: maximum local polynomial degree $p=2$, number of global candidates $n_{c,g}=1000\ M$, number of local candidates $n_{c,l}=5P$, minimum number of samples for local \PCE\ construction $\ns=1.5 P$, minimum number of iterations before checking for restart $n_r=20$, and $\bbeta$ are obtained by LARS and OLS algorithm. Minimum number of samples in sub-domains required to justify an expansions for SSE was set identically to \DALPCE{} and polynomial order is adaptively selected in the range $p\in[2,6]$. Since the SSE is not a~sequential approach, the presented results were obtained for 10 discrete sample sets of increasing size to compare convergence of the method. Note that all samples and candidates are generated by LHS for all compared approaches, though it was shown \cite{NovVorSadShi:CMAME:21} that for the  variance-based sequential sampling, it is significantly better to use advanced techniques such as Coherence D-optimal sampling \cite{CohDOpt}.

\subsection{One-dimensional Toy Example}
\begin{figure}[!ht] 
    \centering
	\includegraphics[width=0.8\textwidth]{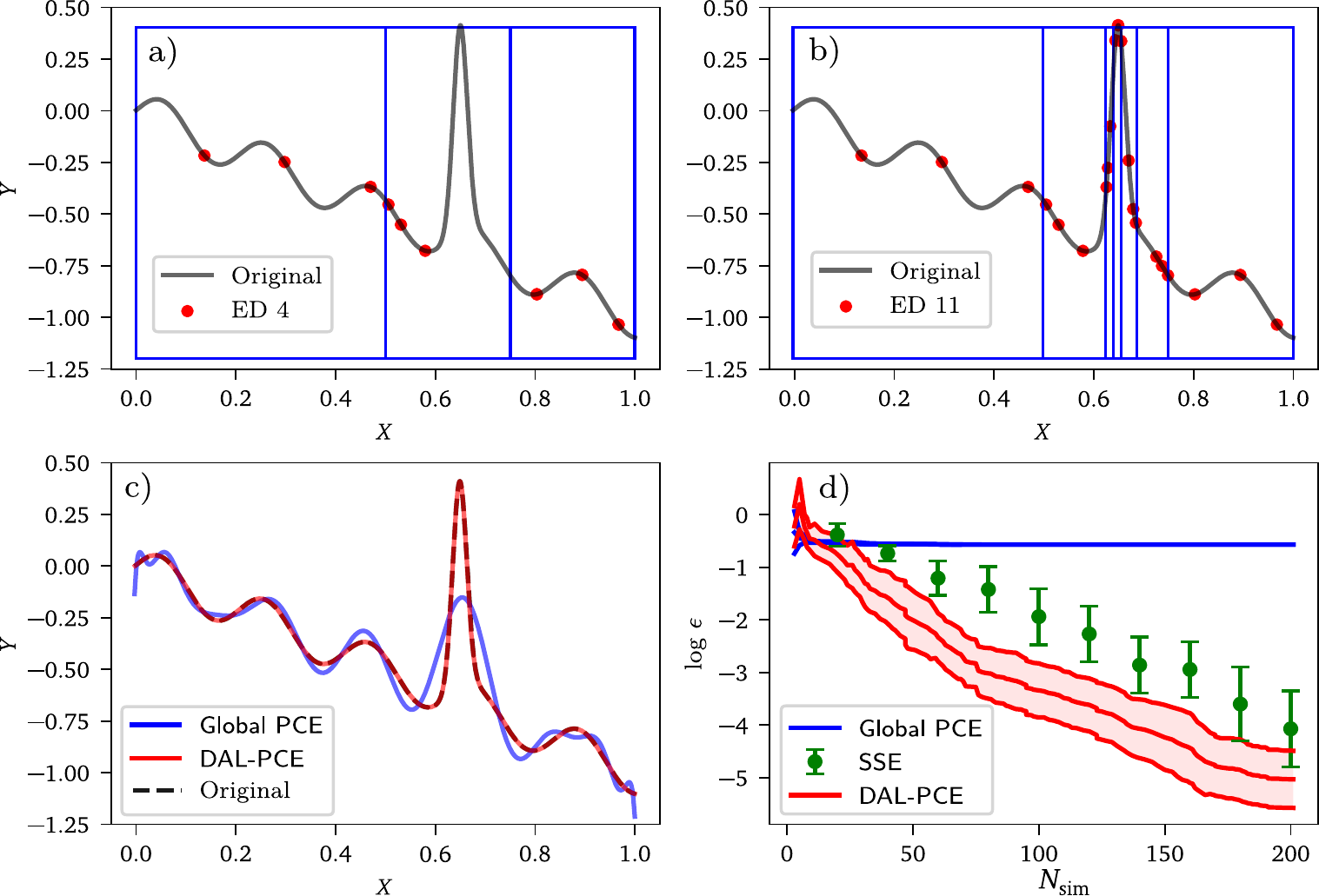}
	\caption{(a), (b) The adapted domain and ED before (iteration 4) and after (iteration 11) exploration and discovery of the exponential part of the mathematical model. (c) Final surrogate models from global \PCE{} and \DALPCE{}. (d) Convergence plot comparing the mean square error for global \PCE\, SSE, and \DALPCE{}. The convergence plots for Global \PCE{} and \DALPCE{} show continuous mean value $\pm \sigma$ intervals from 100 repeated trials, while those for SSE are plotted for several discrete ED sizes.}
	\label{Fig.: 1D function}
\end{figure}

The first example involves a~simple 1D function \cite{SSE:Marelli:21} that is extremely difficult to approximate with \PCE\ due to the third, highly nonlinear ``$\exp$'' term:
    \begin{equation}
  f(X)=-X+0.1 \sin(30X)+\exp(-(50(X-0.65))^2), \quad X\sim \pazocal{U}[0,1]
  .
    \end{equation}
The poor performance of a~single global \PCE\ learned from 200 samples is depicted by the blue line in Fig.~\ref{Fig.: 1D function}c where it is clear that a~single global \PCE\ is not able to accurately approximate the function even for a~high number of samples and high maximum polynomial order $p\in[5,25]$. This function was originally developed to demonstrate the efficiency of SSE based on domain decomposition and thus it is a~natural choice for comparison of the proposed \DALPCE{} and SSE. 

Fig.~\ref{Fig.: 1D function}a-b show a~typical realization of the \DALPCE{} where the algorithm sequentially decomposes the domain and adds additional samples to the ED. Specifically shown are the 4th and 11th iterations. The boundaries of sub-domains are represented by blue vertical lines and red dots show the positions of samples in the ED. Once the algorithm discovers the highly nonlinear region (the steep peak caused by $\exp$), it progressively refines this region and adds more samples there as a~result of the high variance density. 
Of course, these figures show only one realization of the algorithm and the decomposition is dependent on the initial ED. Therefore, it is necessary to repeat the algorithm many times with random initial ED to assess convergence.
Fig.~\ref{Fig.: 1D function}d shows convergence of the error $\epsilon$ from 100 repeated trials. The single global \PCE\ is unable to accurately approximate the original function even when using high $p$ and thus the $\epsilon$ does not converge, as expected. Both methods based on domain decomposition (\DALPCE{} and SSE) achieve great accuracy already for $200$ samples. However, the \DALPCE{} consistently has $1$--$2$ orders of magnitude higher accuracy than SSE for the given number of samples. 
Moreover, increase in variance of $\epsilon$ is, in general, slower in \DALPCE{} than in SSE. Fast increment in variance of SSE can be seen also in the original paper \cite{SSE:Marelli:21}.
Finally, we again observe that convergence is continuous with \DALPCE{}, where convergence can only be assessed at discrete sample sizes with SSE through a~new analysis. 
All of these advantages of the \DALPCE{} can be attributed to the active learning, which both explores the space and exploits the behavior of the function to decompose the domain and add samples. 
Although active learning might lead to lower accuracy (higher $\epsilon$) initially (for small $\ns=10$--$20$) as it is dominated by exploration, it rapidly improves once it identifies important features and begins to favor exploitation.

\subsection{Two-dimensional Singularity}

\begin{figure}[t] 
  \centering
	\includegraphics[width=1\textwidth]{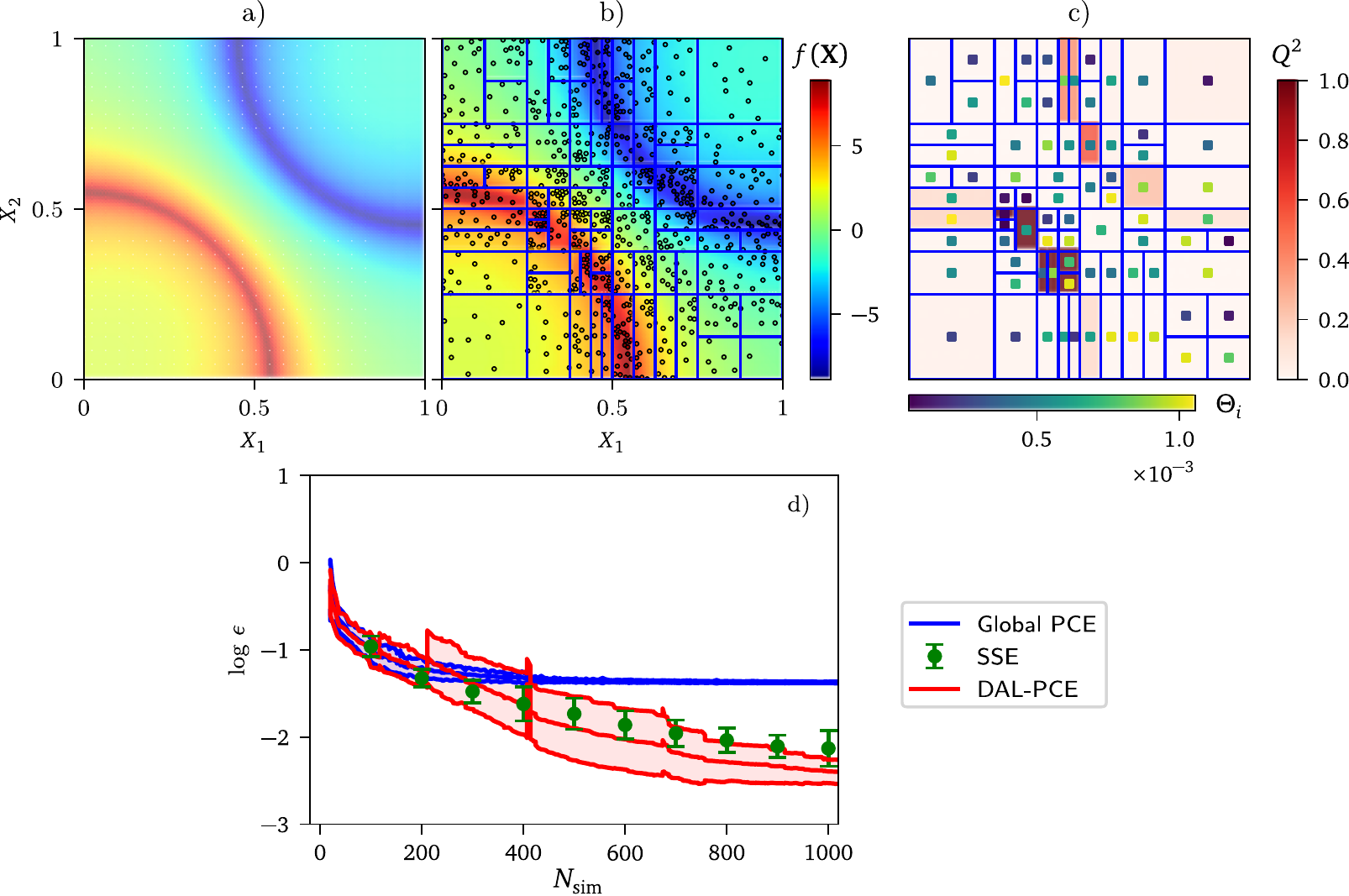}
	\caption{Results for the $2$-dimensional Singularity function: a) original mathematical model, b) approximation via \DALPCE{} (background color), current domain division and the corresponding ED, c) local LOO-CV $Q^2_{\mathcal{D}_i}$ and $\Theta_i$ value for each sub-domain, d) convergence plots for \DALPCE{}, Global \PCE, and SSE showing the mean value and $\pm\sigma$ interval. Convergence plots for SSE show the mean $\pm \sigma$ at discrete sample sizes.}
	\label{Fig: Shields_mirrored}
\end{figure}  

\label{Subsection: 2D singularity Example}
The second example involves a~$2$D function with mirrored quarter-circle arc line singularities \cite{NovVorSadShi:CMAME:21}. The form of the function is give by:
\begin{equation}
\centering
    f(\X)=  \frac{1}{ \lvert 0.3-X_1^2 - X_2^2\rvert + \delta}-
    \frac{1}{ \lvert 0.3-(1-X_1)^2 - (1-X_2)^2\rvert + \delta}, \quad \X \sim \pazocal{U}[0,1]^2
    ,
\end{equation}
where the strength of the singularities is controlled by the parameter $\delta$, which we set as $\delta=0.1$. 
The singularities in this example represent a~challenging task for a~global \PCE\ even with high order, due to the well-known Gibbs phenomenon \cite{davis2022gibbs}. It is thus beneficial to identify the location of the singularity, locally decompose the domain, and construct low-order local \PCE{}s. 

Fig.~\ref{Fig: Shields_mirrored} illustrates the decomposition and \DALPCE{} approximation at a~given stage of the computation. Panel a) visualizes the true values of the function via a~background color. The same coloring scheme is used in panel b) for the pointwise information available in the current ED (small circles) and for the function approximation via \DALPCE{} by the background color. Panels b) and c) show also the final domain decomposition. The symmetry in the decomposition documents the great convergence of the \DALPCE{} thanks to an adaptive decomposition described in the previous section.  Plot c) shows the local $Q^2_{\mathcal{D}_i}$ error in each individual sub-domain (darker color corresponds to higher local error). These local errors clearly show localization of the prediction error to very small areas near singularities, which are continually being refined. The color of the small solid squares in the center of each sub-domains shows the $\Theta_i$ value for that sub-domain.

Finally, the convergence plot in Fig.~\ref{Fig: Shields_mirrored}d) shows that both \DALPCE{} and SSE outperform the global \PCE{}, as expected. The SSE performs comparable to or slightly better than \DALPCE{} for small \Ns, but the \DALPCE{} begins to outperform SSE as \Ns\ grows thanks to the active learning approach that targets samples in the vicinity of the singularities.
Note that the error converges for both SSE and \DALPCE{} as we approach 1000 samples and does not seem to substantially reduce after this. This is due to the fundamental limitation of trying to approximate this singularity, even locally, with low-order polynomials.

\subsection{$M$-dimensional Discontinuity}

\begin{figure}[t] 
  \centering
	\includegraphics[width=1\textwidth]{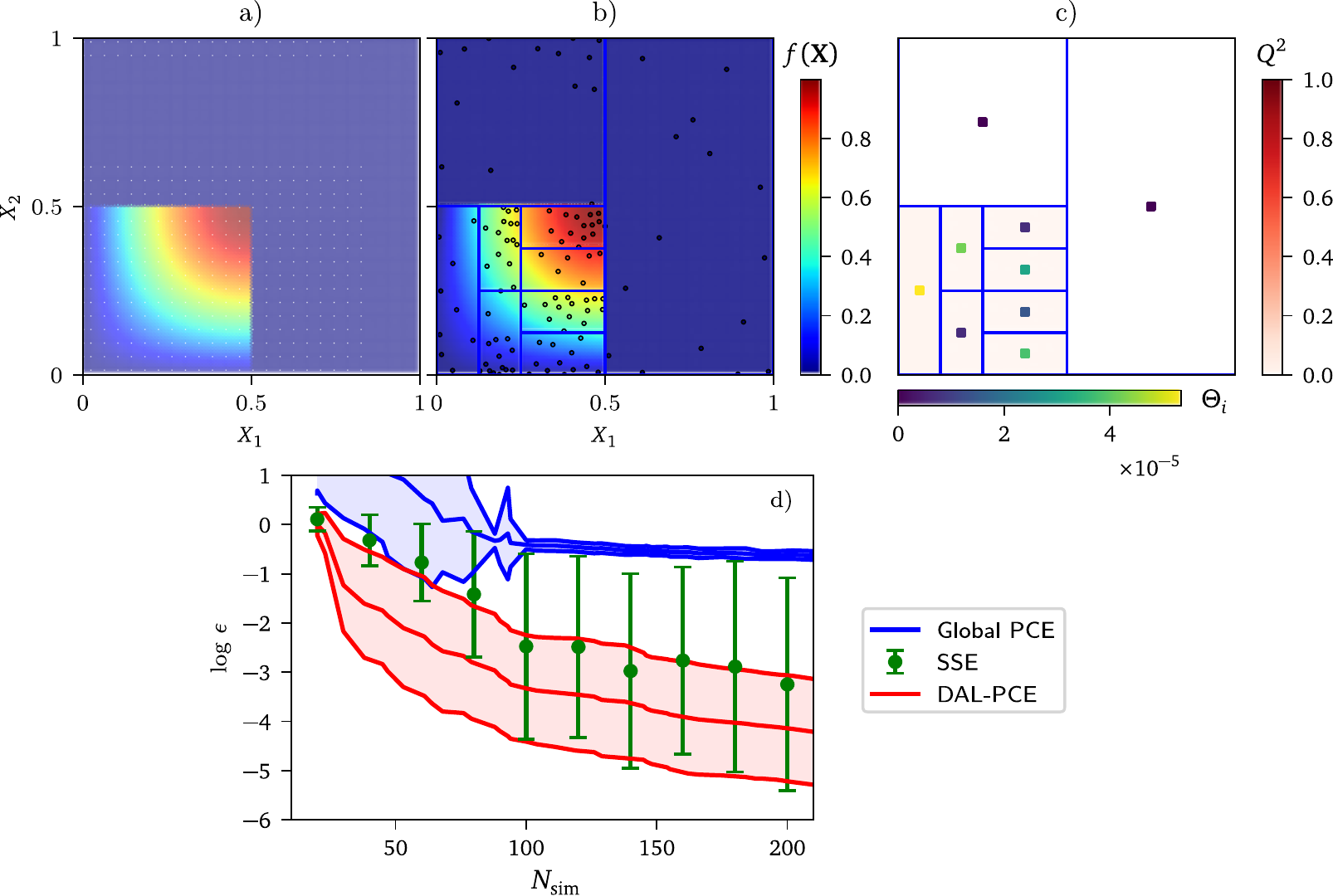}
	\caption{Results for the $2$-dimensional discontinuiy function: a) original mathematical model, b) approximation via \DALPCE{} and ED, c) local LOO-CV $Q^2_{\mathcal{D}_i}$ and $\Theta_i$ value for each sub-domain, d) convergence plots for \DALPCE{}, Global \PCE, and SEE showing the mean value and $\pm\sigma$ interval. Convergence plots for SSE show the mean $\pm \sigma$ at discrete sample sizes.}
	\label{Fig: 2D_disc}
\end{figure}  

The third example investigates the role of dimensionality on the performance of the proposed \DALPCE{}. The following discontinuous function is defined for an arbitrary number of input random variables $M$ \cite{BHADURI2018732}:
\begin{equation}
    f(\X)= 
    \begin{cases}
      \sin{(X_1 \pi)} \sin{(X_2 \pi)} & \text{if $x_1 \leq 0.5 $ and $x_2 \leq 0.5 $} \\
      \sum_{i=3}^M X_i& \text{otherwise}
    \end{cases},
    \quad \X \sim \pazocal{U}[0,1]^M
    .
    \label{eqn:function_3}
\end{equation}
This function has a~discontinuity in the first two input random variables, which can be seen in Fig.~\ref{Fig: 2D_disc}a. A~single global \PCE\ cannot accurately approximate the function because of the discontinuity, although the function $f(\X)$ can be easily approximated by two separate \PCE s in the two regions for which the definitions differ. But, this requires \textit{a priori} knowledge of the discontinuity location.
Since the location of the discontinuity is assumed to be unknown, this function is a~good example for domain adaptation using \DALPCE. 

The detailed results for a~2D version of this problem are depicted in Fig.~\ref{Fig: 2D_disc} in identical form as in the previous example. Note that the local $Q^2_{i}$ errors Fig.~\ref{Fig: 2D_disc}c show perfect accuracy in the part of the input random space where $f(\X)=0$ and thus the associated sub-domains are not preferred for further decomposition. The convergence plot in Fig.~\ref{Fig: 2D_disc}d confirms that a~single global \PCE\ is not able to create an accurate approximation and adding more points to ED does not lead to significant improvements in the approximation. The mean values of errors $\epsilon$ associated to the proposed \DALPCE{} approach are significantly lower in comparison to SSE ($1$--$2$ orders of magnitude) similarly as in the first example, though the convergence trend is similar for both methods. SSE, however, uses a~random splitting routine. This can lead to very high variance of results, since the accuracy is highly dependent on the pattern of the decomposed input random space. This clearly shows the advantage of an active learning approach.

\begin{figure}[t] 
   \centering
	\includegraphics[width=0.9\textwidth]{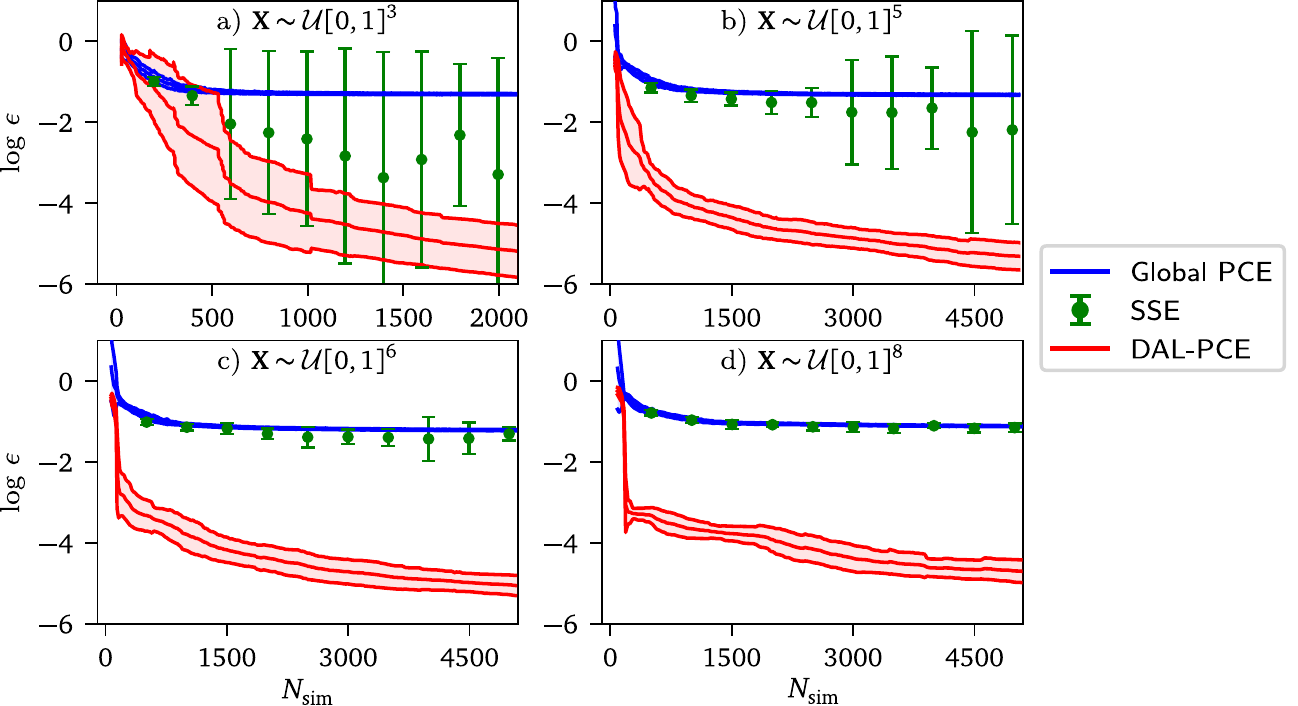}
	\caption{Convergence plots for the $M$-dimensional function: a) $3$-dimensional version, b) $5$-dimensional version, c) $6$-dimensional version, and d) $8$-dimensional version. Convergence plots for the \DALPCE{} and global \PCE\  show the mean value $\pm \sigma$ interval.  Convergence plots for SSE also show the mean $\pm \sigma$, but at discrete sample sizes.}
	\label{Fig: N discontinuity}
\end{figure} 

The influence of dimensionality $M$ on convergence of the \DALPCE{}, SSE, and global \PCE\ is studied in Fig.~\ref{Fig: N discontinuity} for a) 3, b) 5, c) 6, and d) 8 input random variables. 
As the domain dimension increases, the linear part of the function $f(\X)$ occupies an increasing proportion of the domain while the discontinuity remain low-dimensional. The proposed \DALPCE{} greatly improves the convergence because it is able to identify an ideal decomposition and local samples to resolve the discontinuity. For low-dimensions ($M=2,3$), SSE error $\epsilon$ shows a~decreasing trend that is better than global \PCE{} but has an extremely high variance. This is caused by a~lack of control in sample placement. The domain decomposition in SSE is a~product of sample location and without active learning to guide sample placement, SSE will sometimes produce a~very good decomposition and sometimes a~very poor decomposition.
Meanwhile, the proposed \DALPCE{} errors have comparably low variance for low-dimensions and consistently have accuracy comparable to, or better than, the best SSE realizations. 

As the dimension, $M$, increases the \DALPCE{} is able to maintain a~very high level of accuracy, while the accuracy degrades completely for the SSE such that it is comparable to the global \PCE{}. The \DALPCE{} is able to maintain its low error because the discontinuity remains low-dimensional and the active learning process is able to target this region for domain refinement and sampling. This means that the \DALPCE{} remains largely independent of the problem dimension, and instead depends predominantly on the intrinsic dimension of the discontinuous/nonlinear features of the model. The performance of SSE, on the other hand, degrades with dimension because its domain decomposition depends only on a~set of \textit{a priori} specified points that are not selected in a~way that is aware of the important features of the model. Consequently, as the dimension increases the algorithm becomes less likely to refine the domain appropriately around an embedded low-dimensional feature. 
We remark that this desirable scalable convergence trend of the \DALPCE{} is not likely a~universal property, as the trend may break down in problems where the intrinsic dimension of the discontinuity/nonlinearity is high or where the discontinuity occupies a~very small proportion of the domain -- in which case exploration of the space to find the important feature may take a~very large number of samples.

\begin{figure}[t] 
   \centering
	\includegraphics[width=0.9\textwidth]{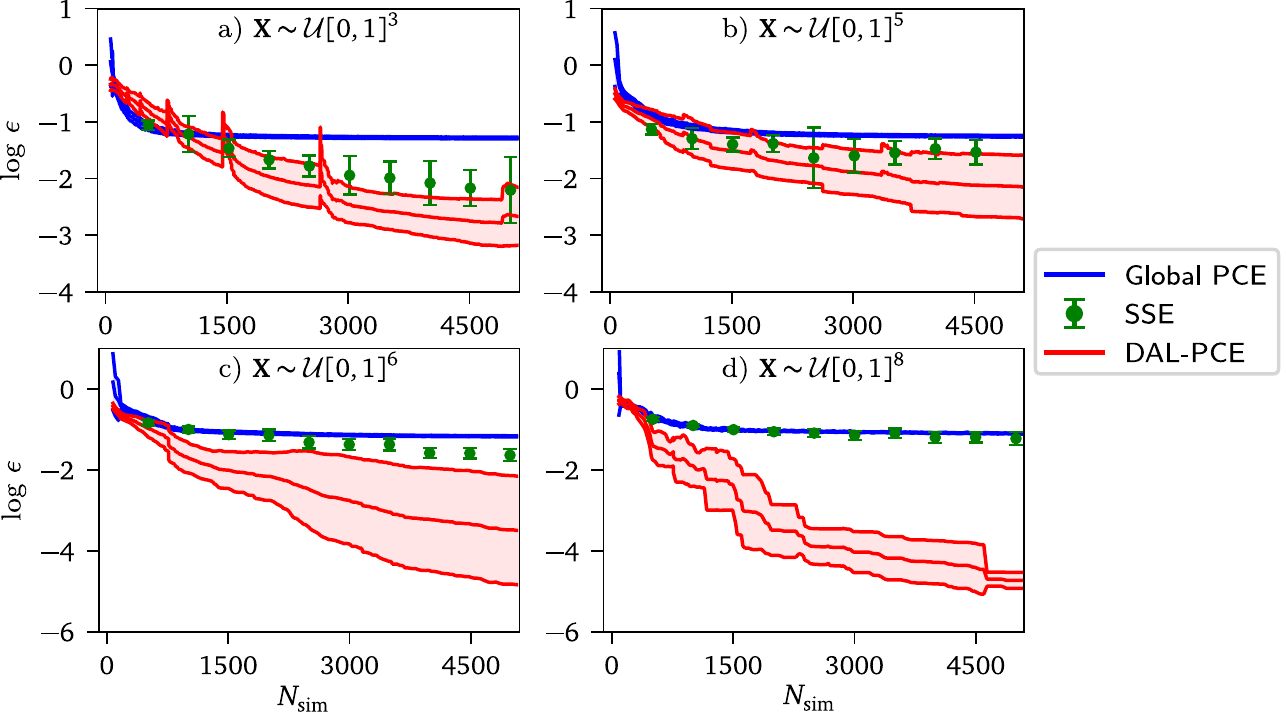}
	\caption{Convergence plots for the modified $M$-dimensional function: a) $3$-dimensional version, b) $5$-dimensional version, c) $6$-dimensional version, and d) $8$-dimensional version. Convergence plots for the \DALPCE{} and global \PCE\  show the mean value $\pm \sigma$ interval.  Convergence plots for SSE also show the mean $\pm \sigma$, but at discrete sample sizes.}
	\label{Fig: 2D discontinuity 061}
\end{figure}

In the present example, the discontinuity in the function given in Eq.\ \eqref{eqn:function_3} lies at $x_1 =0.5 $ and $x_2 = 0.5$, which corresponds to the exact location where the domain will be split for both SSE and during the early iterations of the \DALPCE{}. One might argue that this presents an unreasonable advantage for the proposed algorithm. 
We therefore modified the function such that the discontinuity lies at $x_1 =0.61 $ and $x_2 = 0.61 $. 
Fig.~\ref{Fig: 2D discontinuity 061} shows the convergence for the \DALPCE{} and SSE for this modified function with varying dimension, $M$.
The absolute errors $\epsilon$ exhibit slower decrease, especially for dimensions $M=3$ and $M=5$. However, the proposed active learning still leads to superior results (especially for higher dimensions as in the previous case).  Note that there are visible spikes in the \DALPCE{} convergence graph for the 3-dimensional example. Although the results were statistically processed, these spikes are caused by the restart adaptivity occurring at the same $\Ns$ in each replication. In this case, the optimal decomposition pattern is very complicated and therefore the algorithm activates the restart adaptivity frequently (after multiples of $n_{r}$ steps), until it finds a~suitable pattern to continue convergence. SSE in the 3- and 5-dimensional cases has higher mean error and significantly lower variance in comparison to the previous example. This is caused by the fact that the modified discontinuity location no longer lies along the boundary of the domain decomposition. In the previous example, some SSE realizations achieved near-perfect accuracy because the domain was coincidentally divided along the discontinuity.

\begin{figure}[t] 
   \centering
	\includegraphics[width=0.9\textwidth]{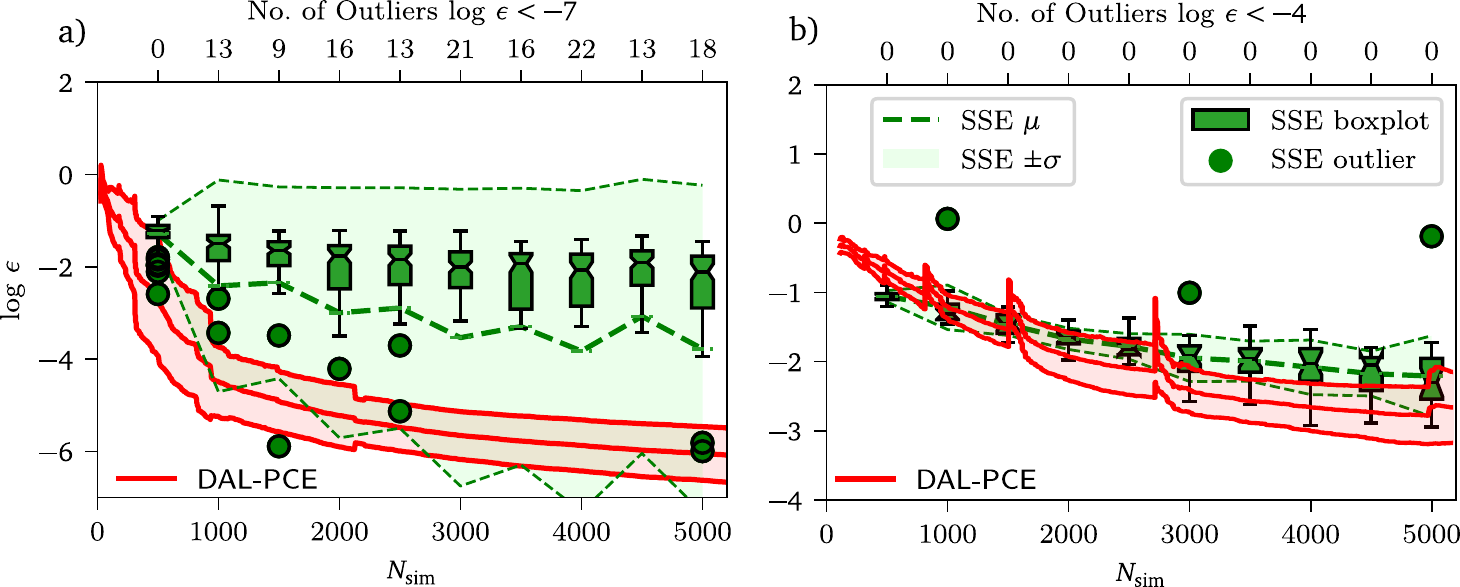}
	\caption{Convergence plots for \DALPCE{} and SSE with additional boxplots for SSE showing the median, lower and upper quartiles and outliers for: a) the 3D example with discontinuity at $x_1 =0.5 $ and $x_2 = 0.5 $, b) the 3D example with discontinuity at $x_1 =0.61 $ and $x_2 = 0.61 $.}
	\label{Fig: SSE outliers}
\end{figure} 

This phenomenon is investigated more closely in Fig. \ref{Fig: SSE outliers}, which compares number of outliers in both versions of 3D examples. In addition to the mean $\pm \sigma$ seen previously, the figure also shows standard boxplots for SSE (median along with lower and upper quartiles) and the corresponding number of ``extreme'' realizations producing very high accuracy (top axis) for a) the original position of discontinuity; and b) discontinuity at $x_1 =0.61 $ and $x_2 = 0.61 $. As can be seen, in panel~a) there are many outliers producing $\epsilon<-7$, which effectively decreases $\mu$ relative to the median while also significantly increasing the variance. In contrast \DALPCE{} has no outliers and it leads to very consistent results. In panel~b), there are no outliers for either SSE or \DALPCE{} and the results are thus consistent with low variance for both methods.

\subsection{Asymmetric shallow von Mises truss}

\begin{figure}[t!] 
   \centering
	\includegraphics[width=1\textwidth]{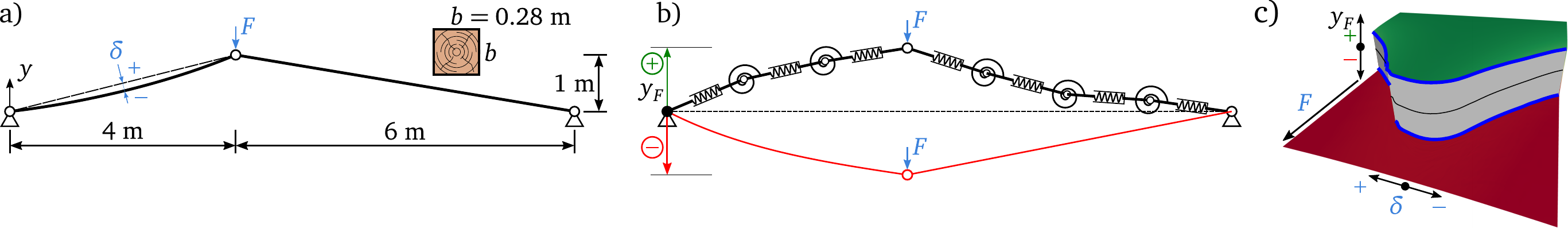}
	\caption{
        Asymmetric shallow von Mises truss. 
        a) Initial geometry with two random variables $F$ and $\delta$;
        b) illustrative sketch of the discrete dynamical model and the meaning of output variable $y_F$, c) illustration of the discontinuous response function of the two input variables.
        }
	\label{fig:vonMisesSketch}
\end{figure}

In this section, we demonstrate the relevance of the proposed method for a~representative engineering example exhibiting discontinuous response. 
Consider the shallow two-bar planar truss subjected to a~vertical load at its top joint, as presented  in \cite{Vor:AdaptPf:22} and illustrated in Fig.~\ref{fig:vonMisesSketch}a.

The truss is formed by two prismatic bars made of a~hard wood (density $800$~kg/m$^3$, modulus of elasticity $E=12$~GPa). There are two variables in the studied von Mises truss: (i) the loading vertical force $F$, and (ii) a~half sine-wave imperfection of the left bar having magnitude $\delta$, see the sketch in Fig.~\ref{fig:vonMisesSketch}a. The load is applied dynamically as a~step function at time zero for an unlimited duration. The structure is modeled, as illustrated in Fig.~\ref{fig:vonMisesSketch}b. In particular, the mass of the bar is concentrated in 21 mass points, including the supports and the loading point. These mass points are connected via $10+10$ translational springs representing the normal stiffness of the true bars. The pairs of the axial members are connected via rotational spring having zero moment for a~zero angle between adjacent bars. The only exceptions are the loading ans support points where there are no rotational springs attached (hinges). The damping is associated with the mass points via linear viscous damping coefficient set to $11~\text{N}\cdot\text{s}/(\text{kg}\cdot\text{m})$ approximating the relative damping of about 3\%.  Explicit dynamics solver \textsf{FyDiK} \cite{fydik,FrantikvonMises} was used to solve the equations of equilibrium at the mass points. The numerical solution lasts to up to two seconds, which is the time needed for almost complete stabilization of the solution (kinetic energy drops below a~negligible threshold).

Since the structure is very shallow, sudden application of the vertical force can cause snap-through buckling, wherein the loading point drops down between the supports and the members switch from a~state of compression to tensile stresses in the final stable state. We specifically study the horizontal coordinate $y_F$ of the loading point after the dynamic response stabilizes to the final deformed shape. The force $F \in (31.6, 772.6)$~kN and initial imperfection $\delta \in (-0.4,0.4)$~m are treated as uniform random variables mapped to the unit square such that the model input $\mathbf{X}\sim \pazocal{U}[0,1]^2$.
Because of the potential snap-through buckling, the solution is discontinuous as illustrated in Fig.~\ref{fig:vonMisesSketch}c.
On each side of the discontinuity, the solution $y_F$ is smooth and slowly-varying having values near +1 m and -1 m, respectively.
Note that the output is \emph{not symmetric} with respect to $\delta=0$ because the dynamical response evolves differently for concave and convex initial displacements.

\begin{figure}[t!] 
   \centering
	\includegraphics[width=1\textwidth]{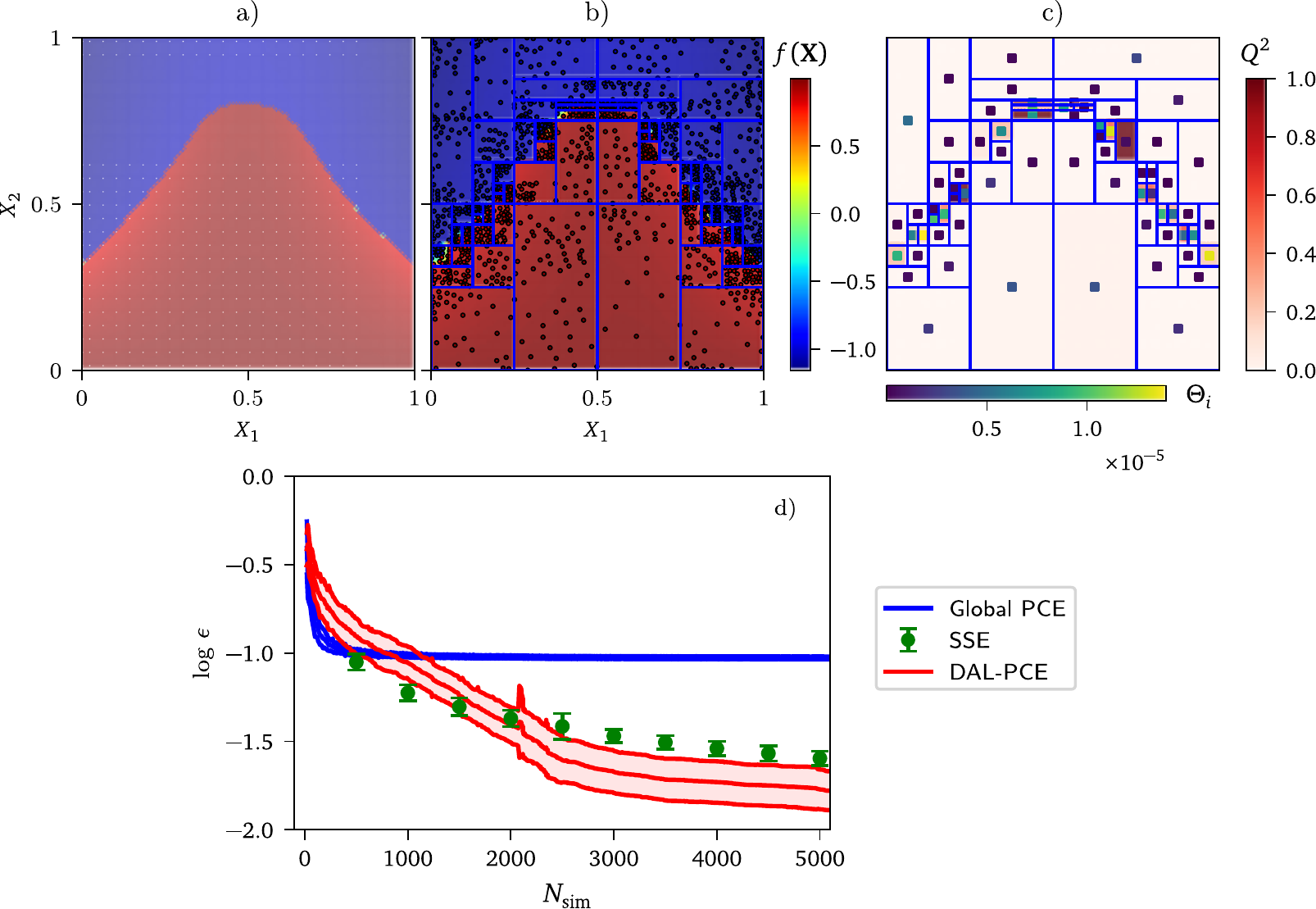}
	\caption{Results for the von Misses truss example: a) original mathematical model (numerical solution), b) approximation via \DALPCE{} and ED, c) local LOO-CV $Q^2_{\mathcal{D}_i}$ and $\Theta_i$ value for each sub-domain, d) convergence plots for \DALPCE{}, Global \PCE, and SSE showing the mean value and $\pm\sigma$ interval; convergence plots for SSE show the mean $\pm \sigma$ at discrete sample sizes.}
	\label{Fig: VonMisses_results}
\end{figure}

The sharp boundary between the buckled and unbuckled regions, shown in Fig. \ref{Fig: VonMisses_results}a cause global \PCE{} to produce poor approximations that are vulnerable to the Gibbs phenomenon, similar to the example in subsection \ref{Subsection: 2D singularity Example}. This is shown by the convergence plots in Fig.~\ref{Fig: VonMisses_results}d comparing global PCE, \DALPCE{}, and SSE. Clearly, the complexity of this example and the complicated shape of the discontinuity limits the accuracy of all the surrogate models. The proposed \DALPCE{} achieves low accuracy for small sample sizes because the corresponding small number of sub-domains and low-order \PCE{}s are unable to sufficiently approximate the boundary. Therefore, the global \PCE{} and SSE (with a~low number of embedding levels) are initially better. With increasing number of samples, the proposed \DALPCE{} approach leads to superior results because the active learning is able to resolve the discontinuity as illustrated in Fig.~\ref{Fig: VonMisses_results}b, which shows the domain decomposition and approximation after 2000 samples. Fig.~\ref{Fig: VonMisses_results}c shows the corresponding LOO-CV errors for each subdomain, demonstrating the errors are confined to small, localized regions near the boundary.

\newpage
\section{Discussion \& Future Work}
The proposed \DALPCE{} approach is a~general methodology for the decomposition of the input random space and construction of localized \PCE{}s using active learning. The proposed active learning is based on a~novel $\Theta$ criterion that optimally balances global \emph{exploration} with local \emph{exploitation} of the model. Although this paper presents one specific learning algorithm, the methodology is general and amenable to modifications to reflect the specific user's needs. The whole process can be divided into two tasks: A) decomposition of the input random space and B) construction of localized \PCE s; and both can be easily modified as discussed further:

\begin{enumerate}[label=\Alph*)]
\item The most important sub-domain $\mathcal{D}_i$ is identified by extended $\Theta$ according to Eq.~\eqref{Eq:ModifiedThetaCrit} evaluated for a~large number of global candidates. In this paper, we use standard LHS for candidate generation, but it may be beneficial to use different sampling methods that produce more uniform coverage of the whole input random space (see e.g.\ \cite{VorEli:Technometrics:20,VorMasEli:Nbody:ADES:19,EliVorSad:miniMax:ADES:20}). Although it is generally possible to generate a~large number of candidates, it might be challenging to uniformly cover the entire input random space, especially in high dimensions. Thus, one can use any sampling technique suitable for a~specific example, e.g. \cite{VorMas:Nbody2:ADES:20}.

Once the $\mathcal{D}_i$ is identified via Eq.~\eqref{Eq:ModifiedThetaCrit}, it is either divided (providing it contains enough ED points) or the sample is extended inside it, to achieve a~better \PCE{} approximation.
The simplest division occurs by splitting the volume into two parts of identical hypervolume in the direction of the highest first-order Sobol' index. However, the algorithm can accommodate various different approaches. For example, it is possible to divide the $\mathcal{D}_i$ into a~higher number of sub-domains, not just two. Moreover, instead of splitting the domain into parts of equal hypervolume, other criteria can be used. For example, the cutting plane can be positioned so to split the domain variance into equal parts.

\item The user can choose to
employ any existing method to construct the non-intrusive \PCE{}s, including various sparse solvers or adaptive algorithms, which may be preferable for certain applications \cite{LuthenReview}. For example, we use LARS with OLS. However, it is generally more efficient to use active learning based on the $\Theta$ criterion for \PCE{} as shown in  \cite{NovVorSadShi:CMAME:21}, which employs variance-based sequential sampling. This improvement can be integrated within the \DALPCE{} to make local \PCE{} more efficient in each subdomain, and thereby improving the overall convergence. The can be compounded by the use of advanced sampling techniques within the subdomains such as Coherence D-optimal sampling \cite{CoherenceOptPCE,CohDOpt}.
\end{enumerate}

As seen from the previous paragraphs, the whole algorithm can be adapted for specific needs reflecting the characteristics of a~given mathematical model, such as dimensionality, sparsity, non-linearity etc., by simply exchanging components of the proposed algorithm for suitable existing (or new) techniques. Note that even after the modification, the whole methodology based on $\Theta$ criterion is still valid and can be used for uncertainty quantification and surrogate modelling as described in this paper. Moreover, in comparison to SSE, the \DALPCE{} sequentially adds points and divides the sub-domains one-by-one based on information obtained from the previous iteration.

Another significant advantage of the \DALPCE{} is that it provides estimates of the local errors, $Q_{\mathcal{D}_i}$, associated with each sub-domain. Since localized \PCE s are constructed independently, local errors estimate the local accuracy of the surrogate model directly, and can be assembled to provide global error measures. Naturally, local accuracy is very important information that can be used for further probabilistic analysis and active learning. Although this paper does not propose any specific approach for further processing of this information, it could serve as a~main ingredient for various active learning algorithms. For example, it could be directly used to predict uncertainty in industrial applications and possibly extend the ED in a~sub-domain of interest.

Finally, an important topic of further research is to study the behavior of the proposed criterion in higher dimensions. In particular, the geometrical terms $l_{c,s}^M$ and $\mathcal{W}_i$ likely cause poor convergence in high dimensions. Although some preliminary results focused on investigating of $l_{c,s}^M$ in high dimensions was previously performed in the paper \cite{NovVorSadShi:CMAME:21} proposing the original $\Theta$ criterion, it is still necessary to perform an extensive study of its behavior as well as investigating the influence of $\mathcal{W}_i$, which may need to be reformulated for high dimensions.

\section{Conclusion}
The paper presented a~novel approach, domain adaptively localzed \PCE{}, for the adaptive sequential construction of localized \PCE s based on active learning and decomposition of the input random space. It combines adaptive sequential sampling based on the recently proposed $\Theta$ criterion to maintain the balance between exploration of the input random space and exploitation of the current characteristics of the \PCE{} together with the adaptive sequential decomposition of the input random space creating sub-domains approximated by local surrogate models. The methodology offers a~general technique that can be easily adapted or modified for specific functions extending its applicability. The performance of the proposed methodology was validated on several numerical examples of increasing complexity investigating different aspects of the algorithm and leading to superior results in comparison to a~single global \PCE{} and the recently proposed SSE.

\section*{Acknowledgments}
 The first author acknowledge financial support provided by the Czech Science Foundation under project number 22-00774S. 
 Additionally, the major part of this research was conducted during the research stay of the first author at Johns Hopkins University supported by the project International Mobility of Researchers of Brno University of Technology, Czechia under project No.~EF18\_053/0016962.

\bibliographystyle{elsarticle-num}
\bibliography{literatura}

\end{document}